\documentclass{article}
\usepackage[linesnumbered,ruled,vlined]{algorithm2e}
\usepackage{amsmath,amsfonts}
\usepackage{algorithm2e}
\usepackage{array}
\usepackage[caption=false,font=normalsize,labelfont=sf,textfont=sf]{subfig}
\usepackage{textcomp}
\usepackage{stfloats}
\usepackage{url}
\usepackage{verbatim}
\usepackage{graphicx}
\usepackage{cite}
\usepackage{amssymb}

\usepackage[numbers]{natbib}

\usepackage{graphicx}
\usepackage{booktabs}
\usepackage{multirow}
\usepackage{graphicx}
\usepackage{amsmath}
\usepackage{amsfonts}
\usepackage{mathrsfs}  
\usepackage{hyperref} 
\usepackage[capitalize,noabbrev]{cleveref} 
\usepackage{tabularx}
\usepackage{multirow}
\usepackage{caption}
\usepackage{subcaption}
\usepackage{adjustbox}
\usepackage{makecell}
\usepackage{threeparttable}

\usepackage{algorithmic}
\usepackage{array}
\usepackage{xspace}
\usepackage{verbatim} 
\usepackage{amsmath}
\usepackage{nicefrac}
\usepackage{comment}
\usepackage[table,xcdraw]{xcolor} 
\usepackage{dsfont}
\usepackage{xr}
\usepackage{subfiles}
\usepackage{enumitem}
\usepackage{mdframed}
\usepackage{tcolorbox}
\usepackage[T1]{fontenc}
\usepackage{float}

\newtheorem{definition}{Definition}

\DeclareMathOperator{\diam}{diam}

\newcommand{\stdv}[1]{\scriptsize$\pm$#1}

\usepackage{newfloat}
\usepackage{listings}
\DeclareCaptionStyle{ruled}{labelfont=normalfont,labelsep=colon,strut=off} 
\floatstyle{ruled}
\newfloat{listing}{tb}{lst}{}
\floatname{listing}{Listing}
\usepackage{arxiv}

\usepackage[utf8]{inputenc} 
\usepackage[T1]{fontenc}    
\usepackage{url}            
\usepackage{booktabs}       
\usepackage{amsfonts}       
\usepackage{nicefrac}       
\usepackage{microtype}      
\usepackage{lipsum}
\usepackage{graphicx}
\graphicspath{ {./images/} }

\title{From Sparse to Dense: Toddler-inspired Reward Transition in Goal-Oriented Reinforcement Learning}

\author{
 Junseok Park \\
  Seoul National University \\
  Seoul, South Korea \\
  \texttt{jspark@bi.snu.ac.kr} \\
  \And
 Hyeonseo Yang \\
  Seoul National University \\
  Seoul, South Korea \\
  \texttt{hsyang@bi.snu.ac.kr} \\
  \And
 Min Whoo Lee \\
  Seoul National University \\
  Seoul, South Korea \\
  \texttt{mwlee@bi.snu.ac.kr} \\
  \And
 Won-Seok Choi \\
  Seoul National University \\
  Seoul, South Korea \\
  \texttt{wchoi@bi.snu.ac.kr} \\
  \And
 Minsu Lee\footnotemark[1]\\
  Sungshin Women's University \\
  Seoul, South Korea \\
  \texttt{mslee@bi.snu.ac.kr} \\
  \And
 Byoung-Tak Zhang\footnotemark[1] \\
  Seoul National University, AIIS \\
  Seoul, South Korea \\
  \texttt{btzhang@bi.snu.ac.kr} \\
}

\begin{document}
\maketitle
\footnotetext[1]{Corresponding author.}
\begin{abstract}
Reinforcement learning (RL) agents often face challenges in balancing exploration and exploitation, particularly in environments where sparse or dense rewards bias learning. Biological systems, such as human toddlers, naturally navigate this balance by transitioning from free exploration with sparse rewards to goal-directed behavior guided by increasingly dense rewards. Inspired by this natural progression, we investigate the \textbf{Toddler-Inspired Reward Transition} in goal-oriented RL tasks. Our study focuses on \textit{transitioning from sparse to potential-based dense (S2D) rewards} while preserving optimal strategies. Through experiments on dynamic robotic arm manipulation and egocentric 3D navigation tasks, we demonstrate that effective S2D reward transitions significantly enhance learning performance and sample efficiency. Additionally, using a Cross-Density Visualizer, we show that S2D transitions smooth the policy loss landscape, resulting in wider minima that improve generalization in RL models. In addition, we reinterpret Tolman’s maze experiments, underscoring the critical role of early free exploratory learning in the context of S2D rewards. 


\end{abstract}

\section{Introduction}
Reinforcement Learning (RL) is a branch of machine learning where agents make decisions to maximize environmental rewards, balancing between \textit{exploration} -- trying new actions -- and \textit{exploitation}, using known actions to optimize rewards. Adjusting the density of the reward function—between sparse and dense—plays a crucial role in achieving an effective balance, as it directly shapes the agent's exploration and decision-making process~\cite{ibrahim2024comprehensive, hare2019dealing}. However, excessively sparse or dense rewards can bias this balance, hindering effective learning, especially in complex environments with high-dimensional inputs such as egocentric raw image observations from 3D real-world-like settings~\cite{schulman2015high, lomonaco2020continual, xiao2020fresh}.

Therefore, achieving this balance necessitates a deeper understanding of the interplay between sparse and dense reward structures. Sparse rewards, typically provided only upon achieving specific goals, encourage extensive environmental exploration but can significantly slow down learning~\cite{andrychowicz2020learning, knox2023reward}. Conversely, dense rewards offer frequent feedback, accelerating learning but may cause agents to prioritize short-term gains over long-term strategies~\cite{laud2004theory}. Given these trade-offs, relying solely on one type of reward structure may fail to capture the complexities required for effective RL learning.

\begin{figure*}[t!]
\centering
\includegraphics[width=0.9\textwidth]{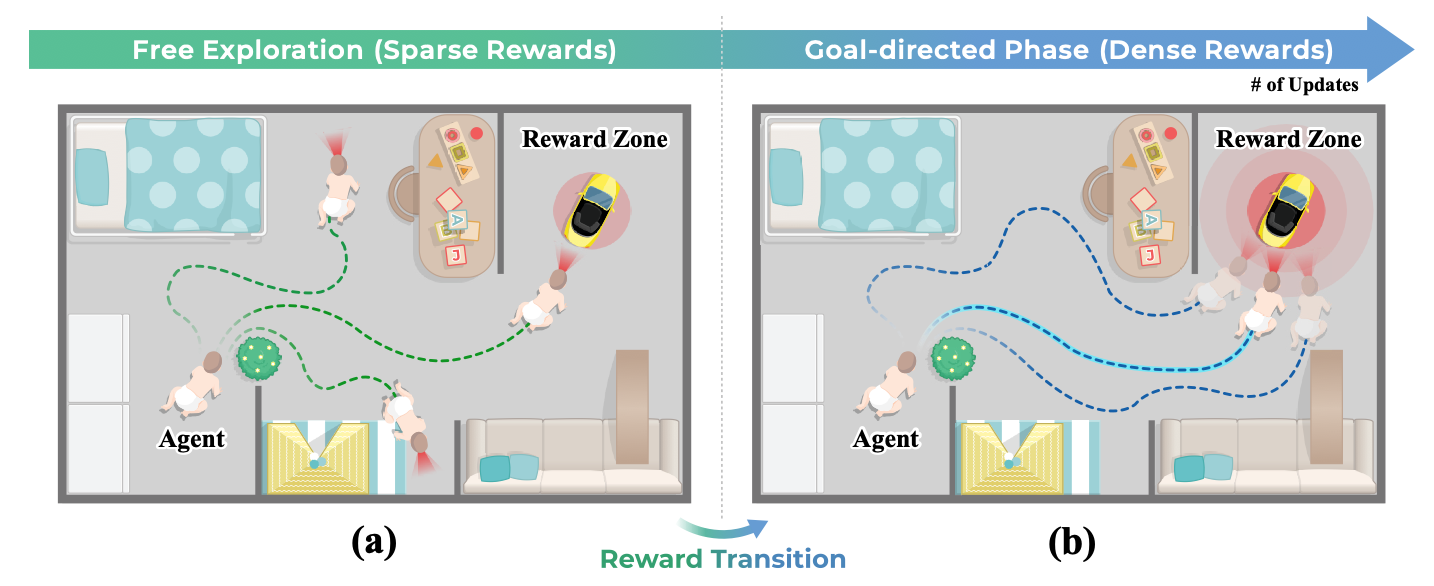}
\caption{Analogy of agents’ trajectories to toddlers’ learning. (a) A toddler’s learning trajectory––free exploration of the environment reflects learning with sparse rewards, (b) goal-directed behavior emerges as the toddler focuses on specific objectives, representing dense rewards. Similarly, the arrow above illustrates the agent’s transition from sparse to potential-based dense rewards, drawing a parallel between the learning processes of toddlers and agents.
}
\label{mainfig}
\end{figure*}

To address this challenge, we draw inspiration from toddlers, who naturally leverage both sparse and dense rewards during their developmental learning processes. Initially, as depicted in Figure~\ref{mainfig}-(a), toddlers act as \textit{innate explorers}, engaging with their environment without prior knowledge—much like agents encountering new situations without expecting immediate rewards~\cite{oudeyer2016evolution}. As they grow, toddlers transition from free exploration to goal-directed learning, focusing on specific objectives with denser rewards, such as visual cues or feedback, as illustrated in Figure~\ref{mainfig}-(b)~\cite{gopnik1999scientist, Gibson, Piaget, gopnik2017changes}. This natural progression provides a compelling analogy for RL dynamics, where agents could similarly refine their strategies through iterative interactions with their environment.


Building on studies of exploration mechanisms in toddlers, we explore this potential of the \textbf{Toddler-inspired Sparse-to-Dense (S2D) Reward Shift} and demonstrate its effectiveness within an RL framework by examining its impact on three key aspects: (1) performance, (2) policy losslandscape, and (3) the role of early free exploration under sparse rewards. For our comparative analysis, we focus on the combination of sparse and dense rewards by evaluating four extrinsic reward strategies that use distance-based cues to achieve the goal: only sparse, only dense, sparse-to-dense, and dense-to-sparse (D2S). To adjust the reward density while maintaining the optimal policy, we incorporate a potential function~\cite{ng1999policy}, an auxiliary reward that guides the agent through changes in the reward structure. Additionally, we leverage intrinsic motivation algorithms~\cite{badia2020never, aubret2019survey, pathak2017curiosity}, which address the exploration-exploitation trade-off by encouraging exploration without explicit external goals, as additional baselines. Performance results indicate that S2D transitions achieve higher success rates and greater sample efficiency compared to other reward strategies in complex goal-oriented RL environments.



To comprehensively assess the impact of S2D transitions on policy learning parameters, we visualize these parameters as a topographical map. In this visualization, each point represents a unique set of parameters, and the altitude corresponds to the policy loss~\cite{li2018visualizing}. Rugged landscapes, characterized by sharp peaks and deep valleys, indicate volatile and challenging learning dynamics, whereas smoother terrains suggest more stable and efficient optimization processes. Our findings reveal that the Sparse-to-Dense (S2D) Reward Transition markedly smooths the loss landscape, as illustrated in Figure~\ref{LossLandscape}. Especially, smoother loss landscapes are associated with wider minima, enhancing generalization by yielding solutions that are less sensitive to minor variations in parameters or data~\cite{keskar2017on}. Furthermore, we use a sharpness metric~\cite{foret2021sharpnessaware} to confirm that S2D results in the widest minima in neural networks after training, outperforming other reward baselines, as shown in Table~\ref{table:sharpness}.


To deepen our understanding of the role of early sparse rewards in facilitating free exploration within the S2D framework, we take inspiration from the work of Edward C. Tolman, a cognitive psychologist, whose maze experiments~\cite{tolman1948cognitive} demonstrated the concept of latent learning—an implicit process in which initial free exploratory behavior enables the formation of a cognitive map of the environment before the introduction of explicit rewards. To reinterpret this in RL frameworks, we designed two egocentric 3D maze environments, where randomized goal and spawn locations enhance generalization, and enriched visual stimuli encourage agents to learn diverse object representations. Analogously, our experimental results indicate that early free exploration during the sparse reward phase in the S2D framework allows agents to establish robust initial parameters, as shown in Figure~\ref{figure:analysis} and Figure~\ref{figure:trajectory}. These parameters could, in turn, enhance the generalization and stability of policy optimization during the subsequent dense reward phase.



Our research sheds light on the intricate balance between exploration and exploitation in RL, providing key insights for designing adaptive reward structures. To support these findings, we developed diverse testbeds, including dynamic robotic arm manipulation and egocentric 3D navigation tasks, specifically designed to evaluate and enhance generalization. By drawing inspiration from toddlers' natural learning behaviors, we bridge biological and artificial learning, providing a fundamental groundwork for RL systems that are not only robust and generalizable but also efficient in complex environments.

This study builds upon our earlier work~\cite{park2024unveiling} and offers the following key contributions:

\begin{enumerate}

\item \textbf{Performance Improvement:}  
Inspired by toddler learning patterns, we demonstrate that the S2D approach effectively enhances RL learning by balancing exploration and exploitation, leading to higher success rates, improved sample efficiency, and better generalization compared to other reward strategies.


\item \textbf{Validation Across Diverse Environments:}
We validate our approach for generalization and robustness across diverse environments, including manipulation and visual navigation tasks. To this end, we also designed customized 3D environments, such as ViZDoom and Minecraft mazes, for comprehensive evaluation.

\item \textbf{Impact on 3D Policy Loss Landscape:}  
Using a cross-density visualizer and sharpness metric, we show that S2D transitions smooth the policy loss landscape, resulting in wider minima that improve generalization in RL policies.

\item \textbf{Reinterpretation of Tolman’s Maze Experiment:}  
We show that the role of early free exploration under sparse rewards in S2D frameworks establishes robust initial policies, enhancing generalization and stability during transitions to dense rewards.

\end{enumerate}

\section{Related Works}

\subsection{Exploration-Exploitation in Deep Reinforcement Learning}

Balancing exploration and exploitation is a key challenge in deep RL~\cite{ladosz2022exploration}. Exploration allows agents to discover new strategies, while exploitation maximizes rewards from known behaviors. Striking this balance is particularly challenging in sparse-reward settings, where feedback is rare and tied to specific goals, offering little guidance for effective learning. To address this, additional rewards are introduced through two complementary methods. Extrinsic rewards, aligned with task objectives, provide feedback for intermediate milestones, guiding agents toward their goals. Intrinsic rewards, driven by curiosity or novelty, promote exploration of new states using techniques like next-state prediction~\cite{badia2020never, aubret2019survey, pathak2017curiosity}. These mechanisms work together to help agents overcome the limitations of sparse rewards by encouraging exploration while maintaining goal-oriented behavior. Within this framework, we propose a reward strategy inspired by human development. Similar to toddlers, who initially explore freely in sparse-reward environments before transitioning to goal-directed behaviors supported by denser feedback, we investigate how this paradigm can enhance RL agents’ adaptability, exploration efficiency, and overall performance across varying reward structures.

\subsection{Toddler-Inspired Learning}
The developmental stages of toddlers have provided a novel perspective for advancing deep learning. Researchers studied the natural exploratory behaviors and unique learning mechanisms of toddlers and discovered ways to refine both supervised and reinforcement learning approaches. For example, classifiers trained on datasets reflecting a toddler’s perspective of objects have been shown to outperform those based on adult perspectives~\cite{bambach2018toddler}, demonstrating the benefits of exploration-centered learning. Similarly, critical learning periods in toddlers correspond to similar phases in RL ~\cite{park2021toddler, de2022critical} and deep neural networks~\cite{achille2018critical}. These toddler-inspired methodologies highlight significant parallels between biological growth and AI model development, underscoring the value of biological insights in advancing AI.

\subsection{Curriculum Learning}

Curriculum Learning (CL), inspired by educational curriculums, has been shown to improve training speed~\cite{Hacohen2019OnTP}, learning efficiency, and safety~\cite{Turchetta2020SafeRL} in machine learning. The progression of CL from easy to more challenging tasks is effective in enhancing generalization and convergence rates~\cite{Bengio2009CurriculumL, Weinshall2018CurriculumLB} in both supervised and reinforcement learning~\cite{florensa2018automatic, graves2017automated, narvekar2020generalizing}. While numerous studies focus on easy-to-hard tasks ~\cite{kalAntidis20hard, du21self, dong2017class}, other studies~\cite{zhang1994selecting, mackay1992information} suggest a \textit{general-to-specific} approach. In such an approach, agents first gather diverse experiences and then exploiting them. Following this idea, we incorporate the toddler-inspired S2D reward transition into RL, applying it to goal-directed reward transitions.

\subsection{Potential-Based Reward Shaping (PBRS)}

In RL, the objective is to maximize cumulative rewards. However, designing optimal reward functions often poses significant challenges, frequently involving intensive reward engineering. Reward Shaping (RS) is a well-established method used to accelerate training by offering supplementary feedback~\cite{taylor2009transfer}. When reward structures are variable, potential-based reward shaping ensures that optimal strategies remain stable by integrating rewards based on potential functions~\cite{ng1999policy}. Traditionally, these shaped rewards are applied consistently throughout the training process. In contrast, our study introduces the concept of Toddler-Inspired Reward Transition, examining the impact of dynamically adjusting reward density over time.

\section{Preliminaries}
\subsection{Reinforcement Learning} 
Reinforcement learning (RL) is a field of machine learning particularly suited for solving sequential decision-making problems. The core principle of RL is to maximize an agent's expected reward through trial and error, analogous to how humans acquire skills to complete tasks. RL problems are commonly modeled using a Markov Decision Process (MDP), defined as $\langle \mathcal{S}, \mathcal{A}, \mathcal{P}, \mathcal{R}, \gamma\rangle$, which consists of the following components: a set of states $\mathcal{S}$, a set of actions $\mathcal{A}$, a state transition probability matrix $\mathcal{P}: \mathcal{S} \times \mathcal{A} \rightarrow \mathcal{S}$, and a reward function $\mathcal{R}: \mathcal{S} \times \mathcal{A} \rightarrow \mathbb{R}$. The discount factor $\gamma$ is used to limit the influence of rewards from distant future states in a trajectory.

At each time step $t$, the agent selects an action $a_t \in \mathcal{A}$ based on a policy $\pi(\cdot|s_t)$, which specifies a probability over actions given the current state $s_t \in \mathcal{S}$. The MDP updates its state to $s_{t+1} \sim \mathcal{P}(\cdot|s_t,a_t)$, and the agent receives a reward $\mathcal{R}(s_t,a_t)$ during the transition. The goal of an RL algorithm is to determine an optimal policy $\pi^*\in \Pi^* \subseteq \Pi$, where $\Pi$ is the set of all possible policies, and $\Pi^*$ represents the subset of policies that maximizes the expected cumulative reward $R = \mathbb{E}\left[\sum_{t=0}^{\infty} \gamma^{t}\mathcal{R}\left(s_{t}, a_{t}\right)\right]$.

\begin{figure*}[t!]
\centering
\includegraphics[width=0.9\textwidth]{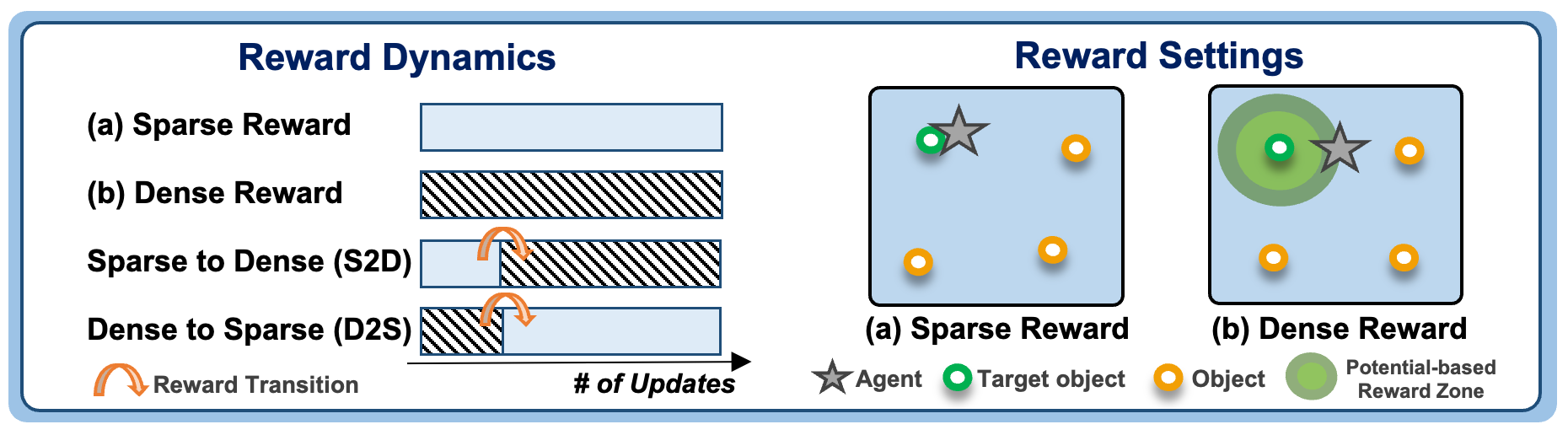}
\caption{Summary of the baseline rewards.
}
\label{Reward Baseline}
\end{figure*}

\subsection{Potential-Based Reward Shaping}

To improve an agent’s performance, selecting an appropriate curriculum is crucial. In this study, we argue that adjusting the proportions of provided rewards is instrumental in achieving robust and generalized performance. Formally, we define $\text{supp}(\mathcal{R}) \subseteq \mathcal{S}$ as the \textit{support} set of the reward function $\mathcal{R}$. In other words, $\text{supp}(\mathcal{R})$ comprises the states that yield non-zero rewards for certain actions:
\begin{equation*}
\text{supp}(\mathcal{R})=\{s\in \mathcal{S} \mid \exists a\in \mathcal{A} \,\,\, s.t. \,\,\, \mathcal{R}(s, a) \neq 0 \}.
\end{equation*}
The sparsity of a reward function is quantified by the ratio of the cardinalities of $\text{supp}(\mathcal{R})$ and $\mathcal{S}$. For two reward functions $\mathcal{R}_{D}$ and $\mathcal{R}_{S}$ defined on $\mathcal{S}$, we say that $\mathcal{R}_{D}$ is \textit{denser} than $\mathcal{R}_{S}$ if the condition $|\text{supp}(\mathcal{R}_{S})|\le|\text{supp}(\mathcal{R}_{D})|$ is satisfied.
In the context of curriculum learning, we assume that the support set of a \textit{dense} reward function $\mathcal{R}_{D}$ encompasses that of a \textit{sparse} reward function $\mathcal{R}_{S}$: $\text{supp}(\mathcal{R}_{S})\subseteq \text{supp}(\mathcal{R}_{D})$.

For reward transition, mechanisms that systematically move from sparse to dense rewards while maintaining learning stability are essential. Potential-based reward shaping (PBRS) provides a practical approach by densifying the reward signal with an additional potential-based reward $F_i$, all while preserving the optimal policy. In PBRS, the potential-based reward $F_{i}$ for the $i$-th MDP is defined as follows:

\begin{equation}
F_i(s, a) = \mathbb{E}_{s' \sim \mathcal{P}(s, a)}[\gamma \Phi_{i}(s') - \Phi_{i} (s)],
\end{equation}
where $\Phi_i : \mathcal{S} \rightarrow \mathbb{R}$ is a \emph{potential function} at stage $i$. Note that the optimal policy $\pi^*\in\Pi^*$ with respect to reward $\mathcal{R}_{i}$ is still optimal with respect to reward $(\mathcal{R}_{i}+F_i)$:
\small
\begin{align*}
Q(s, a) &= \mathbb{E}_{\mathcal{P}, \pi}\left[\sum_{t=0}^{\infty} \gamma^t\left(\mathcal{R}^t_{i}+F^t_{i}\right) \mid s_0=s\right] \\
&= \mathbb{E}_{\mathcal{P}, \pi}\left[\sum_{t=0}^{\infty} \gamma^t\mathcal{R}^t_{i} 
+ \gamma^t \big(\gamma \Phi_{i}(s_{t+1}) - \Phi_{i}(s_t)\big) \mid s_0=s\right] \\
&= \mathbb{E}_{\mathcal{P}, \pi}\left[\sum_{t=0}^{\infty} \gamma^t \mathcal{R}^t_{i}\right] - \Phi_i(s_0).
\end{align*}
Also, the supported region of the PBRS reward, denoted as $\text{supp}(\mathcal{R}_{i}+F_i)$, contains the region of its original reward $\mathcal{R}_{i}$:
\begin{equation*}
\text{supp}(\mathcal{R}_{i}+F_i)=\text{supp}(\mathcal{R}_{i})\cup\text{supp}(F_i)\supseteq\text{supp}(\mathcal{R}_{i}),
\end{equation*}
and it means that the PBRS reward is more denser than the original reward.

\subsection{Multi-stage RL with Potential-based Reward Function}
\label{sec:currl}
Curriculum learning~\cite{Hacohen2019OnTP, Turchetta2020SafeRL} is a multi-stage approach for  training models robustly by progressively adjusting the difficulty of tasks over time. In RL, curriculum learning is defined as a series of MDPs $\{\mathcal{M}_i\}_{i=1}^{N}$ where each MDP $\mathcal{M}_i = \langle \mathcal{S}, \mathcal{A}, \mathcal{P}, \mathcal{R}_i, \gamma \rangle$ is characterized by a unique reward function $\mathcal{R}_i$, representing different task difficulties~\cite{Bengio2009CurriculumL, Weinshall2018CurriculumLB}. By setting the stage transitions $\mathcal{T}=(T_1, T_2,\cdots,T_{N-1})$, the MDP transitions from one to another.

\begin{definition}[Curriculum]
\label{def:curriculum}
Let a series of MDPs be $\{\mathcal{M}_i\}_{i=1}^{N}$ with $\mathcal{M}_i = \langle \mathcal{S}, \mathcal{A}, \mathcal{P}, \mathcal{R}_i, \gamma \rangle$, and its state transitions be $\mathcal{T} = (T_1, T_2,\cdots,T_{N-1}) \in \mathbb{N}^{N-1}$. A \textit{curriculum} $\mathscr{C}$ is defined as a tuple $\mathscr{C}=(\{\mathcal{M}_i\}_{i=1}^N, \mathcal{T})$ where the $\mathcal{M}_{I(t;\mathcal{T})}$ is chosen to train the agent at training step $t\in\mathbb{N}$. The stage indicator $I(t;\mathcal{T})$ is defined as:
\[
    \forall i,\forall t \in \left[T_{i-1}, T_i\right), \quad I(t;\mathcal{T}) := i,
\]
\end{definition}
where $T_0:=0$ and $T_N:=\infty$.


\subsection{Wide Minima Phenomenon and Loss Landscape}

In deep neural networks, the loss landscape refers to the multi-dimensional space where each point’s altitude represents the loss for specific parameters~\cite{li2018visualizing}.  The objective is to find minima in this landscape. Wide minima have broad gradients, facilitating smooth convergence to global minima via gradient descent, which enhances robustness and generalization to data distribution perturbations~\cite{keskar2016large}. In contrast, \textit{sharp minima} possess steep gradients that are sensitive to such perturbations, often leading to overfitting and poor generalization~\cite{goodfellow2014qualitatively}. Empirical studies have shown that models located in wide minima tend to perform better and generalize more effectively than those situated in sharp minima~\cite{keskar2017on, j2018finding}. This principle also applies to RL, where the distribution of the experiences of an agent can vary slightly at each time step. Our empirical results confirm that policies positioned in wide minima improve generalization and robustness in these fluctuating environments.


\subsection{Tolman’s Maze Experiment}

The classic maze experiment conducted by Edward C. Tolman provides a foundational basis for understanding the Sparse-to-Dense (S2D) reward transition strategy~\cite{tolman1948cognitive}. Tolman’s study revealed how rats navigated mazes under varying reward conditions, yielding valuable insights into the role of free exploration and reward timing. Specifically, three groups of rats were tested:

\begin{enumerate}
    \item No Reward Group: Rats freely explored the maze without receiving any rewards. (analogous to \textbf{only sparse})
    \item Consistent Reward Group: Rats received rewards consistently upon reaching the goal. (analogous to \textbf{only dense})
    \item Delayed Reward Group: Rats began in a reward-free phase but later transitioned to consistent rewards. (analogous to \textbf{Sparse-to-Dense, S2D})
\end{enumerate}

Notably, the Delayed Reward Group outperformed others once rewards were introduced, suggesting that the period of free exploration allowed the rats to form internal representations, or cognitive maps, of their environment. These cognitive maps facilitated efficient navigation when the rewards became available. Inspired by these cognitive and developmental phenomena, our study explores whether free exploration under sparse rewards in the S2D framework can similarly cultivate foundational experiences in AI agents, thereby enhancing their ability to construct cognitive maps and ultimately improving learning efficiency and policy robustness in RL.





\section{Method}
To implement our experiments, we design a reward transition frameworks inspired by toddler behavior. We investigate how this transition affects agent learning, focusing on its impact on the policy loss landscape and the emergence of wide minima. Inspired by Tolman’s experiments, we further examine the role of free exploration under sparse rewards within S2D frameworks by analyzing the internal representations formed.

\subsection{Toddler-Inspired Sparse to Dense Reward Curriculum}

We first design the Sparse to Dense (S2D) reward transition to infuse the exploration-to-exploitation strategy into curriculum learning. A curriculum $\mathscr{C}$ becomes an \textit{S2D-curriculum} if the reward functions $\{\mathcal{R}_i\}_{i=1}^{N}$ of their respective MDPs $\{\mathcal{M}_i\}_{i=1}^{N}$ gradually become denser while preserving optimal policies. 

\begin{definition}[\textit{Toddler-inspired S2D-curriculum}]
\label{def:Anti_curriculum}
A curriculum $\mathscr{C} = (\{\mathcal{M}_i\}_{i=1}^N, \mathcal{T})$ with its corresponding MDPs $\mathcal{M}_i=\langle \mathcal{S}, \mathcal{A}, \mathcal{P}, \mathcal{R}_i, \gamma \rangle$ is an \textit{S2D-curriculum} if the following conditions are satisfied:
\begin{equation}
\mathrm{supp}(\mathcal{R}_1)\subseteq \mathrm{supp}(\mathcal{R}_2) \subseteq \cdots \subseteq \mathrm{supp}(\mathcal{R}_N) 
\label{cond1}
\end{equation}
\begin{equation}
\Pi^*_1\supseteq \Pi^*_2 \supseteq \cdots \supseteq \Pi^*_N, 
\label{cond2}
\end{equation}
\end{definition}
$\Pi_{i}^*$ is a set of optimal policies within the MDP $\mathcal{M}_i$. Equation \ref{cond1} indicates that the sequence of reward functions should increase in density. Equation \ref{cond2} constrains the optimality on the policies such that the optimal policies of $\mathcal{M}_{i+1}$ are also optimal in $\mathcal{M}_{i}$.

From Equations~\ref{cond1} and~\ref{cond2}, the reward functions must become denser while preserving the same set of optimal policies. To achieve this, we use the potential-based reward shaping (PBRS) approach \cite{Ng1999PolicyIU, harutyunyan2015expressing}, which allows adjusting the reward density without altering the optimal policy. 


For the experiments, we assume that the agent can only get a reward if it reaches the goal $g\in\mathcal{G}$ within a certain radius in the \textit{sparse} reward setting ($\mathcal{M}_1$): $F_1(s)=0$. On the other hand, in the \textit{dense} reward setting ($\mathcal{M}_2$, $\mathcal{M}_3$), the agent gets an additional potential-based dense reward $F_{i\ge2}$ with the potential function $\Phi(\cdot)$, shown in Equation~\ref{eq:dense_guidance}:
\begin{equation}
    \Phi(s) := \diam_p(\mathcal{S}) - ||s - g||_p,
    \label{eq:dense_guidance}
\end{equation}
where $s \in \mathcal{S}$ and $g \in \mathcal{G}$ are the agent's current position and the goal position, respectively. $\text{diam}_p(\mathcal{S})$ is the diameter of given set $\mathcal{S}$. The dense reward is determined by the agent's proximity to the goal, based on the Euclidean distance ($p = 2 $) or Manhattan distance ($p = 1$) . Table~\ref{tab:reward-formulations} shows sparse and dense reward functions utilized across various experimental environments. The detailed implementation of the Toddler-Inspired S2D Reward Transition is provided in Algorithm~\ref{app:algorithm}.


\begin{footnotesize}
\begin{algorithm}[H]

\KwIn{RL algorithm $\mathcal{G}$ (e.g., SAC, PPO, DQN), Curriculum $\mathscr{C}=\{\mathcal{M}_k\}_{k=1}^{n}$ with state transition $\mathcal{T}=\{T_k\}_{k=1}^{n-1}$, Potential function $\Phi$, Discount factor $\gamma$, Terminal step $T_{d}$}
\KwOut{Trained RL agent with optimized policy $\pi_\theta$}
\SetAlgoLined

Initialize RL agent with policy parameters $\theta$, environment $\mathcal{E}$

Initialize Replay buffer $\mathcal{B}\gets\emptyset$

$T\leftarrow0, k \leftarrow 1$

\While{$T<T_{d}$}{
    $t \leftarrow 0$
    
    Reset environment, obtain $s_0$
    
    \While{not terminal condition}{
        $t \leftarrow t + 1$
    
        $a_t \sim \pi_\theta(\cdot|s_t)$

        $(r_t,s_{t+1})\leftarrow\mathcal{M}_k(s_t,a_t)$
        
        \textcolor{gray}{\# $F_1(\cdot,\cdot)=0$, $\text{supp}(F_k)\subseteq \text{supp}(F_{k+1})$}
    
        $F_k(s_t, a_t) \gets \gamma \Phi(s_{t+1}) - \Phi(s_t)$
        
        $\tilde{r}_t \gets r_t + F_k(s_t, a_t)$ \quad \textcolor{gray}{\# Update reward}
                
        $\mathcal{B}\leftarrow \mathcal{B} \cup \{(s_t, a_t, \tilde{r}_t, s_{t+1})\}$

        $b\leftarrow \text{sample}(\mathcal{B})$

        $\pi_{\theta}\leftarrow \mathcal{G}(\pi_{\theta},b)$ \quad \textcolor{gray}{\# Update policy with mini-batch from replay buffer}
    }
    $T\leftarrow T + t$
    
    \If{$T \geq T_k$}{
        $k \gets k + 1$ \quad \textcolor{gray}{\# Transition to next stage}
    }
}
\Return{$\pi_\theta$}
\caption{Algorithm for Toddler-Inspired Sparse-to-Dense (S2D) Reward Transition in RL}
\label{app:algorithm}
\end{algorithm}
\end{footnotesize}

\subsection{Visualizing Policy Loss Landscapes}

This study examines the impact of the S2D transition on the policy loss landscape. Following the method outlined in \cite{li2018visualizing}, we plot policy loss landscapes by varying parameters $\tilde{\theta} = \theta + \alpha \mathbf{x} + \beta \mathbf{y}$, where $\theta$ denotes the current parameters and $\alpha$ and $\beta$ are normalized coordinates. The axes, represented by vectors $\mathbf{x}$ and $\mathbf{y}$, introduce specific perturbations in the parameter space. These vectors are normalized to have unit length and are orthogonalized for clarity and consistency in scaling. The z-axis represents the average policy loss over a batch of transitions from the replay buffer. It is important to note that the relative position of one landscape over another is not significant since each landscape corresponds to distinct network parameters with different loss ranges due to varying stages of learning.

Given the lack of effective visualization techniques for policy loss landscapes during reward transitions in previous research, we have created the Cross-Density Visualizer. This tool provides a 3D view of the shift of policy loss landscapes from exclusively sparse or dense rewards to mixed-reward settings. Our approach involves two distinct sets of transitions: Sparse-to-Dense (S2D) and Sparse-to-Sparse (Only Sparse) in one, and Dense-to-Sparse (D2S) and Dense-to-Dense (Only Dense) in the other. As illustrated in Figure~\ref{LossLandscape} and further elaborated in Appendices B and C, our visualizations reveal \textit{smoothing effects}, especially prominent in the S2D model.

\begin{table}[H]
    \caption{
    Sparse and dense reward formulations used in each environment. Rewards are provided when the specified conditions are met.
    }
    \centering
    \small  
    \begin{adjustbox}{width=\columnwidth}  
        \begin{tabular}{@{}llcc@{}}
            \toprule
            \rowcolor[HTML]{EEEEEE}
            \textbf{Environment}       & \textbf{Description}         & \textbf{Sparse Reward}          & \textbf{Dense Reward} \\ \midrule
            \textbf{LunarLander}       & 2D landing simulation        & $||s-g||_2 < 1$                 & $\gamma \Phi(s_{t+1}) - \Phi(s_t) < 0.3$ \\ 
            \textbf{CartPole}          & Pole balancing               & $||s-g||_2 < 0.02$              & $\gamma \Phi(s_{t+1}) - \Phi(s_t) < 1$   \\ 
            \textbf{UR5}               & Robotic arm reaching         & $||s-g||_2 < 0.02$              & $\gamma \Phi(s_{t+1}) - \Phi(s_t) < 1$   \\ 
            \textbf{ViZDoom-Seen}      & First-person maze (trained)  & $||s-g||_2 < 0.0075$            & $\gamma \Phi(s_{t+1}) - \Phi(s_t) < 0.14$ \\ 
            \textbf{ViZDoom-Unseen}    & First-person maze (unseen)   & $||s-g||_2 < 0.0075$            & $\gamma \Phi(s_{t+1}) - \Phi(s_t) < 0.14$ \\ 
            \textbf{Cross Maze}        & 2D navigation task           & $||s-g||_1 < 2$                 & $\gamma \Phi(s_{t+1}) - \Phi(s_t) < 5$   \\ 
            \textbf{Playroom Maze}     & 3D toddler exploration       & $||s-g||_1 < 2$                 & $\gamma \Phi(s_{t+1}) - \Phi(s_t) < 5$   \\ \bottomrule
        \end{tabular}
    \end{adjustbox}
    \label{tab:reward-formulations}
\end{table}

\subsection{Exploring Minima Sharpness After Reward Transitions}

Our findings of smoothing effects prompted us to hypothesize that the S2D transition helps escape local minima and enhances generalization in wider minima. Wide minima in neural networks are indicative of robust and adaptable models~\cite{keskar2017on, j2018finding}. By investigating minima after this transition, we aim to enhance performance and gain a better understanding of agent adaptability in various situations. To evaluate the extent to which the policy remains in wide minima, we measure the end-of-training convergence of the neural network of S2D to wide minima using the sharpness metric defined in Equation~\ref{defn:sharpness} and compare it with those of other transitions. This follows the approach proposed in \cite{foret2021sharpnessaware}, which outlines a specific form of sharpness measure as described in \cite{keskar2017on}.

\begin{equation}
    \max_{||\epsilon||_p\leq \rho} L_{\pi}(\theta+\epsilon)-L_{\pi}(\theta)
    \label{defn:sharpness}
\end{equation}
Here, $\theta$ represents the current parameters in the policy loss landscape. The maximizer $\hat{\epsilon}$ can be estimated using the following equation:
\begin{equation*}
\hat{\epsilon}=\rho \,\text{sgn}(\nabla_\theta L_{\pi}(\theta))\cdot \nicefrac{|\nabla_\theta L_{\pi}(\theta)|^{q-1}}{\big( ||\nabla_\theta L_{\pi}(\theta)||^q_q\big)^{\frac{1}{p}}},
\end{equation*}
where $1/p + 1/q = 1$, and $\text{sgn}(\cdot)$ is the element-wise sign function~\cite{foret2021sharpnessaware}. For our experiments, we used $\rho=0.02$, and $p=q=2$ in our experiments. 

\subsection{Analyzing Policy Behavior in Tolman’s Maze Experiments}

We also examine the impact of the S2D transition on agents’ internal representations, inspired by Tolman’s maze experiments~\cite{tolman1948cognitive}. We hypothesize that early free exploration under sparse rewards fosters robust initial parameters through diverse experiences, enabling efficient policy learning under dense rewards. Using (1) RNN feature convergence, (2) policy visualization and (3) Visualization of Trajectory, we highlight another dimension of the S2D approach's advantages.

\subsubsection{Measuring the Mean Distance Between RNN Features}
\label{subsubsec:meanDistance}
In our partially observable 3D egocentric playroom maze, the agent uses a recurrent neural network (RNN) to maintain hidden states. To measure how quickly these internal representations converge, we:
\begin{enumerate}[label=(\alph*)]
    \item Collect observations: We fix a particular roll-out (trajectory) in the environment. This trajectory is the same across all reward baselines for fair comparison.
    \item Record and extract hidden states: At regular training intervals (e.g., every $X$ steps), we checkpoint the RNN parameters of each agent. Using the saved parameters, we re-run the same trajectory and record the corresponding hidden state vectors ${\mathbf{h}_{t}}$ at each time step~$t$.
    \item Compute mean distances: We compute the mean pairwise distance, which is  $\ell_2$ distance $| \mathbf{h}_{t_1} - \mathbf{h}_{t_2} |_2$, between hidden state vectors across time to quantify how much (and how consistently) the RNN representation changes across training checkpoints.
\end{enumerate}
A sharper drop in these distances typically indicates that the RNN features are converging faster or more stably. As demonstrated in \cref{figure:analysis}-(a), the S2D transition tends to yield faster convergence than other baselines after reward transition.

\subsubsection{Action Frequency Analysis} Figure \ref{figure:analysis}-(b) presents a temporal analysis of discrete action frequencies, such as \textit{move forward}, \textit{turn left}, and \textit{turn right}, over checkpoints saved at regular intervals during training. The sparse-to-dense reward transition occurred at 3 million steps, and the statistical distribution of actions sampled from the policy $\pi(a \mid s)$ was evaluated over three rollout episodes. For each step $t$ within an episode, actions $a_t$ were drawn according to $\pi(a \mid s_t)$, where $s_t$ denotes the observed state at time $t$. The aggregated frequency of an action $a$, denoted as $f_a$, is computed as:

\[
f_a = \frac{1}{n}\sum_{i=1}^{n} \left(\frac{1}{T_{i}}\sum_{t=1}^{T_i} \mathbb{I}(a = a_t^i)\right),
\]

where $T_i$ represents the length of the $i$-th episode, \( N = \sum_{i=1}^{5} T_i \) is the total number of steps across all five episodes, and $\mathbb{I}(a = a_t^i)$ is the indicator function that equals 1 if the action \( a_t^i \) at step \( t \) of episode \( i \) matches \( a \), and 0 otherwise. This analysis provides insights into the temporal preferences of the policy $\pi$ over the course of training.

\subsubsection{Trajectory Visualization} 
The agent's spatial trajectories, defined by its global coordinates, are plotted on a 2D overhead map to visualize navigation patterns. Figure~\ref{figure:toddlerresult} highlights the most frequently traversed paths, identified via visual inspection. These paths represent common strategies employed by the agents to reach the goal, often corresponding to optimal or near-optimal navigation routes. Frequent trajectories are extracted using frequency-based methods, providing insights into the agent's pathfinding efficiency. As shown in Figure~\ref{figure:trajectory}, compared to an only dense—which tends to be biased toward stronger rewards and thus exhibits more “angular” exploration—S2D explores more freely across multiple directions. This broader range of experiences leads to richer, more foundational learning.





\section{Experiment Design}

In this section, we explore in-depth the dynamics of the S2D reward transition compared to multiple reward-driven methods. We discuss its substantial effects across multiple challenging environments, as illustrated in Appendix A. We specifically explored the implications of applying the reward transition to RL by addressing four critical questions:

\begin{tcolorbox}[colframe=black!50, colback=black!5!white, title=Research Questions, fonttitle=\bfseries, coltitle=black, boxrule=0.5mm, arc=0mm, auto outer arc]
    \begin{itemize}[leftmargin=0pt, itemsep=5pt, parsep=5pt]
    \item \textbf{Performance Improvement}: How does the Toddler-inspired S2D reward transition measure against other reward strategies?
    \item \textbf{Impact on 3D Policy Loss Landscape}: What changes occur in the policy loss landscape following the S2D transition?
    \item \textbf{Link Between Wide Minima and Toddler-Inspired Reward Transition}: Does the S2D transition encourage convergence towards wide minima?
    \item \textbf{Reinterpretation of Tolman's Maze Experiment}: How do early sparse rewards influence an agent’s S2D learning compared to other reward strategies?
    \end{itemize}
\end{tcolorbox}

The details of our environments and extensive supplementary experiments are outlined in Appendix A and C.

\begin{figure*}[t!]
\centering
\includegraphics[width=0.9\textwidth]{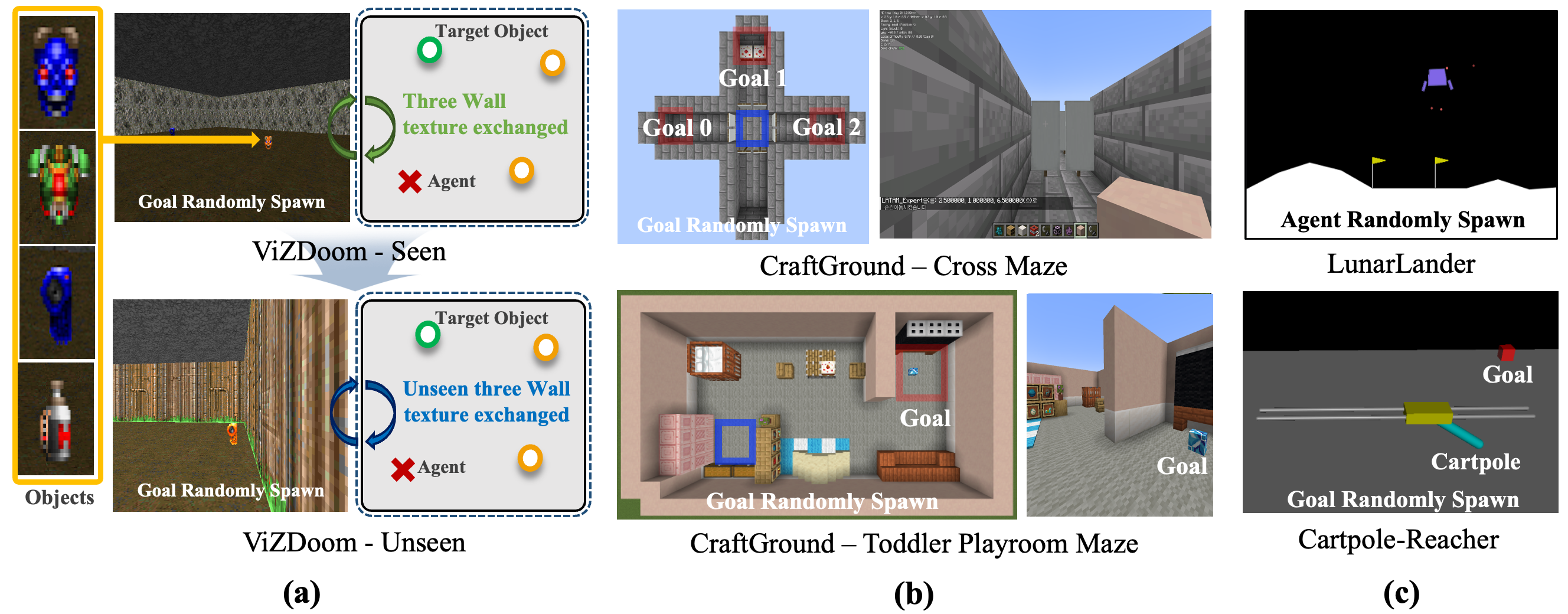}
\caption{Experimental environments. (a) ViZDoom environments. (b) Minecraft environments. (c) Additional environments: Modified UR5-Reacher, Cartpole-Reacher with randomly spawned goals, and the detailed description of LunarLander are provided in Appendix A.
}
\label{Env}
\end{figure*}

\subsection{Reward Setting Details}

\subsubsection{Reward-driven baselines for comparison.} 

As outlined in Figure~\ref{Reward Baseline}, we examine four primary reward settings: (1) Only Sparse, where rewards are given only upon reaching the goal, encouraging broad exploration; (2) Only Dense, which uses potential-based reward shaping (PBRS)~\cite{ng1999policy} to provide additional rewards based on proximity to the goal, preserving optimal policies while enhancing goal-directed learning; (3) Sparse-to-Dense (S2D), starting with sparse rewards to promote exploration before transitioning to dense rewards for effective exploitation; and (4) Dense-to-Sparse (D2S), the reverse of S2D, starting with dense rewards and transitioning to sparse rewards to evaluate its relative impact. For detailed formulations of sparse and dense rewards, including the specific distance thresholds for each environment, please refer to Table~\ref{tab:reward-formulations}.

As an additional baselines, we also incorporate intrinsic motivation reward methods, which similarly tackle this trade-off. Specifically, we used Never Give Up (NGU)~\cite{badia2020never} for discrete environments like ViZDoom and LunarLander, and Random Network Distillation (RND)~\cite{burda2018exploration} for continuous action environments like CartPole. These approaches incentivize exploration by providing intrinsic rewards to agents for identifying new states.

\subsubsection{Hyperparameter Analysis of Reward Transition Timing.}
Furthermore, we analyzed hyperparameters for the timing of reward transitions through ablation studies (see Table~\ref{table:sharpness}). Inspired by early developmental interactions~\cite{Piaget,shonkoff2000neurons}, we compare three transition points, $t \in \{ 1N, 2N, 3N\}$, where $N$ is roughly 1/12 of the entire training period. The specific value of \( N \), adjusted for each environment's episode length, is detailed in Appendix A. These transition points are labeled as $\mathscr{C}_1$, $\mathscr{C}_2$, and $\mathscr{C}_3$, respectively, for S2D and D2S reward transitions.

\subsection{Environment Details}

To evaluate the impact of reward dynamics, we tested under various conditions, including state-based and visual observations, as well as both discrete and continuous action spaces, detailed in Appendix A-Table A.2. We examined different reward configurations, including the S2D reward transition, across several goal-directed tasks in established benchmark environments. Figure~\ref{Env}-(c) depicts examples such as LunarLander~\cite{brockman2016openai}, CartPole, and UR5~\cite{todorov2012mujoco}. Appendix A provides a comprehensive description of the challenging dynamics introduced for UR5 and CartPole, with randomized placements for agents, goals, and obstacles, labeled as the ‘reacher’ version. All agents had full access to state information and were assessed using the Soft Actor-Critic (SAC)~\cite{haarnoja2018soft} algorithm. Additionally, we adjusted the reward structure for both sparse and dense settings, with additional details in Appendix A.

\subsubsection{Enhanced Generalization Environment}
To deepen the evaluation of generalization capabilities, we designed a challenging egocentric navigation scenario within the ViZDoom environment~\cite{kempka2016vizdoom}, as illustrated in Figure~\ref{Env}-(a). In the \textbf{Seen} environment (Appendix Figure A.12-(a), objects were randomly placed, and walls featured one of three textures. The \textbf{Unseen} environment (Appendix Figure A.12-(b)) required the agent to adapt to three new wall textures, distinct from those in the \textbf{Seen} scenario. The A3C~\cite{mnih2016asynchronous} algorithm was employed to assess performance in this context.

\subsubsection{Tolman's Maze Environments}
To emulate the learning behavior in Tolman's maze, we created two 3D egocentric navigation scenarios using the Minecraft toolkit (see Appendix A). The first scenario, illustrated in Figure~\ref{Env}b-Upper, is a cross maze where agents spawn in a designated blue zone at the maze’s center and must move outward along different corridors to reach three goals—two used in training and a newly introduced one for evaluation. The central insignia obscure visibility, encouraging an exploration-exploitation trade-off. This environment is especially challenging for Goal~2 due to its reduced reward area (see Appendix~A, Figure~A.13). 

The second scenario, shown in Figure~\ref{Env}b-Lower, mimics a toddler's playroom, where agents spawn in a blue zone and must navigate to a randomly placed goal in a red zone. The space is cluttered with objects of varying colors and sizes, requiring more complex navigation. This setup is designed to analyze policy representations as agents learn from egocentric observations in a high-dimensional input space that closely resembles real-world conditions.




\begin{figure*}[t!]
     \centering
    \includegraphics[width=0.9\textwidth]{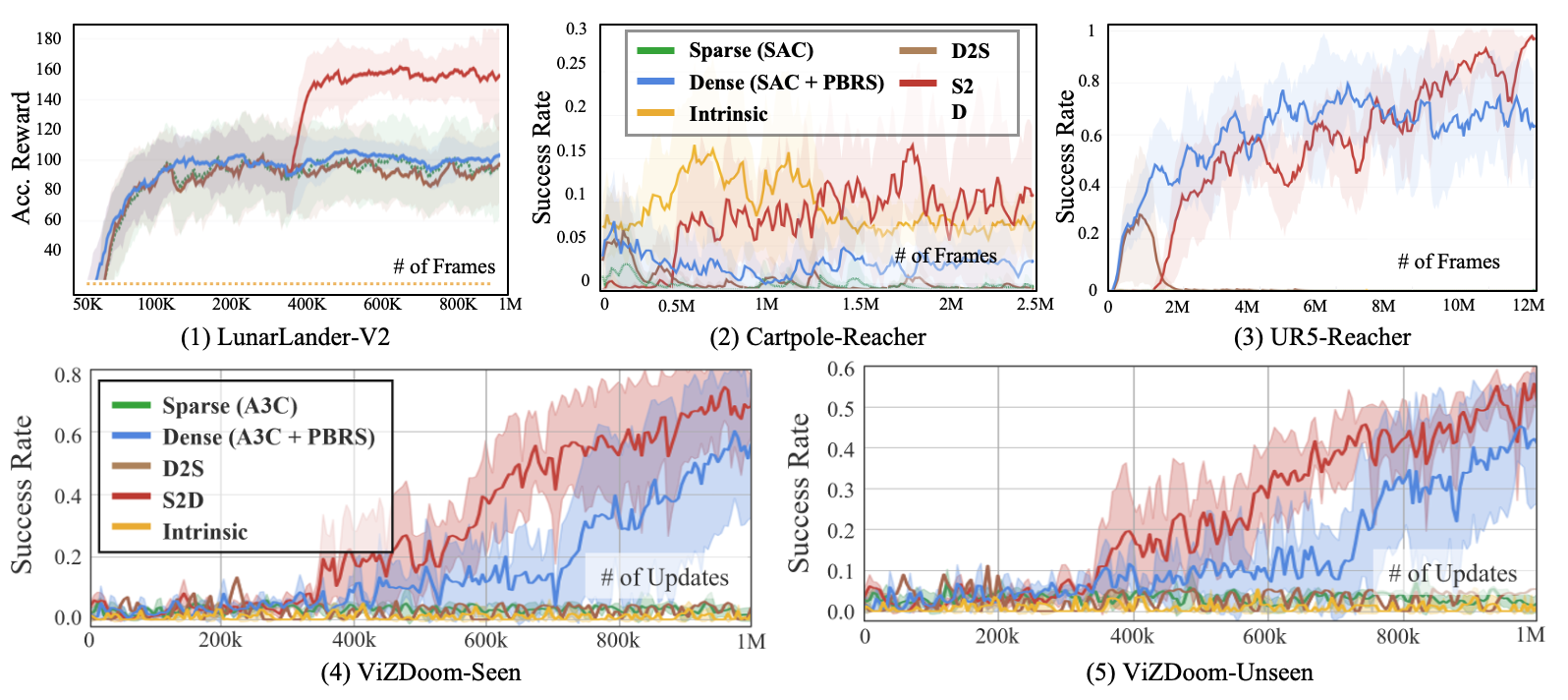}
    \caption{The agent’s performance across different reward baselines in several goal-oriented tasks. (1-3) In LunarLander, the total reward gained from intrinsic incentives was well below zero, as indicated by the dashed line. For UR5, both intrinsic motivation and sparse reward settings resulted in near-zero performance, making it difficult to observe. (4), (5) The ViZDoom agent’s ability to generalize across different reward types. }
    \label{figure:mainresult} 
\end{figure*}

\section{Results}
\subsection{Performance Results}
\subsubsection{Sample Efficiency and Success Rate}

We conducted experiments in diverse environments with static points of view. The results are presented in Figure~\ref{figure:mainresult}-(1-3) and Table~\ref{table:sharpness}. These environments vary in the agents’ performance under sparse reward; LunarLander and CartPole-Reacher exhibited poor performance with default sparse rewards. In these scenarios, the S2D approach consistently outperformed all other baselines and showed superior sample efficiency. Even in the more challenging UR5-Reacher, which requires more precise control and has a higher-dimensional action space, S2D still led the performance. Unlike intrinsic motivation-based algorithms that often prioritize exploration state over goal achievement, S2D outperformed other methods.
Furthermore, we conducted experiments in ViZDoom-Seen and Unseen, Minecraft Cross, and Minecraft playroom maze, which are environments with an egocentric viewpoint. Similar to the results mentioned above, S2D exhibited superior performance across all cases, as demonstrated in  Figure~\ref{figure:mainresult}-(4-5), Figure~\ref{figure:crossmap} and Figure~\ref{figure:toddlerresult}. Notably, D2S outcomes were consistently lower than those of S2D in all environments, highlighting the effectiveness of the S2D transition as a training curriculum.

\subsubsection{Enhanced Generalization Performance} 
The S2D reward transition consistently outperformed other agents in various dynamic environments that require strong generalization, such as those with varying goal locations or agent spawn positions, as shown in Figure~\ref{Env}a to \ref{Env}c. We specifically designed more challenging environments that introduce visual changes not seen during training, illustrated in Figure~\ref{Env}a and \ref{Env}b.

In the ViZDoom-Unseen environment, where agents face significant visual changes due to the addition of three new wall textures (Figure~\ref{Env}a), the S2D transition demonstrates superior generalization and sample efficiency compared to other baselines, as shown in Figure~\ref{figure:mainresult}-(4),(5). Similarly, in the Minecraft Cross maze, where a newly occurring goal location appears during evaluation (Figure~\ref{Env}b-Upper), the S2D transition still displayed superior results, as shown in Figure~\ref{figure:crossmap}.

\begin{table*}[t]
    \centering
    \vspace{0.25em}
    \begin{adjustbox}{width=1\textwidth}
    \begin{tabular}{cccccccccc}
    \toprule
Task  & Metric & S2D($\mathscr{C}_1$)    & S2D($\mathscr{C}_2$)     & \textbf{S2D}($\mathscr{C}_3$)   & Only Sparse & Only Dense & D2S($\mathscr{C}_1$) & D2S($\mathscr{C}_2$) & D2S($\mathscr{C}_3$)  \\
\midrule
\multirow{2}{*}[0.3ex]{\makebox[0pt]{\LARGE \thead{Lunar\\Lander}}}   & Perf. & 138.71\stdv{3.71}          & 63.40\stdv{160.55}   & \textbf{168.88}\stdv{23.66}   & 142.50\stdv{4.25}          & 139.68\stdv{14.90}          &    140.75\stdv{7.46}     & 130.63\stdv{19.69}  &   142.37\stdv{15.62}\\
&  Sharp. & 27.06\stdv{36.31}  & 1231.93\stdv{2424.61}     & \textbf{7.46}\stdv{3.37} & 8.97\stdv{2.83}  & 8.71\stdv{4.43}   &   8.95\stdv{2.89}  & 8.99\stdv{2.97}& 11.32\stdv{3.72}\\
\midrule
\multirow{2}{*}{CartPole}    &    Perf.    & 3.18\stdv{4.00} & \textbf{14.61}\stdv{10.96}   & 5.29\stdv{7.47}  & 0.14\stdv{0.25}  & 3.88\stdv{4.63}  & 1.55\stdv{0.29} & 0.38\stdv{0.07} & 0.97\stdv{0.19}\\
&  Sharp. & 0.12\stdv{0.24}          & \textbf{0.01\stdv{0.15}}          & 0.01\stdv{0.24}          & 0.08\stdv{0.57}        & 0.19\stdv{0.03}          &  0.16\stdv{0.09} & 0.05\stdv{0.21} & 0.02\stdv{0.17} \\
\midrule
\multirow{2}{*}{UR5}    & Perf.    & 65.54\stdv{10.86}  & 65.69\stdv{17.32}      & \textbf{94.15}\stdv{4.28}      & 0.00\stdv{0.00}          & 64.23\stdv{13.03}          & 0.00\stdv{0.00}   & 0.00\stdv{0.00} & 0.00\stdv{0.00} \\
& Sharp. & 0.67\stdv{0.01}          & 0.62\stdv{0.11}          & \textbf{0.61\stdv{0.04}}          & 0.09\stdv{0.52}          & 0.67\stdv{0.01}          & 0.52\stdv{0.24}  & 0.56\stdv{0.28} & 0.47\stdv{0.20} \\
\midrule
Cross Maze 0  & Perf.    & 75.19\stdv{0.06}  & \textbf{81.90}\stdv{0.06}  & 75.49\stdv{0.06}  & 64.94\stdv{0.03}  & 67.32\stdv{0.08}  & 60.57\stdv{0.03}  & 62.88\stdv{0.03}  & 63.96\stdv{0.03} \\
\midrule
Cross Maze 1  & Perf.    & 69.65\stdv{0.07}  & \textbf{77.16}\stdv{0.05}  & 75.39\stdv{0.06}  & 57.82\stdv{0.03}  & 69.46\stdv{0.07}  & 55.66\stdv{0.04}  & 57.31\stdv{0.05}  & 51.95\stdv{0.05} \\
\midrule
Cross Maze 2  & Perf.    & \textbf{63.60}\stdv{0.05}  & 62.86\stdv{0.03}  & 57.60\stdv{0.06}  & 57.26\stdv{0.03}  & 57.54\stdv{0.07}  & 55.15\stdv{0.05}  & 57.25\stdv{0.04}  & 54.47\stdv{0.04} \\
\midrule
playroom maze & Perf.   & 22.95\stdv{0.03}  & 21.94\stdv{0.02}  & \textbf{25.78}\stdv{0.03}  & 17.50\stdv{0.02}  & 18.91\stdv{0.02}  & 16.06\stdv{0.01}  & 17.22\stdv{0.01}  & 17.59\stdv{0.01} \\
\bottomrule
\end{tabular}
\end{adjustbox}
\caption{Performance and sharpness metrics were measured over at least six trials in each environment. Reduced sharpness indicates wide minima, which may improve generalization performance. The best performance and corresponding sharpness values are highlighted in bold, showing that the top-performing \textbf{S2D} also achieves the widest minima. Through ablation studies of the reward transition timing, we also found that the optimal reward transition timing occurs within the first third of training, similar to toddlers' early critical learning period.}
\label{table:sharpness}
\end{table*}

\begin{figure*}[t!]
\centering
\includegraphics[width=0.9\textwidth]{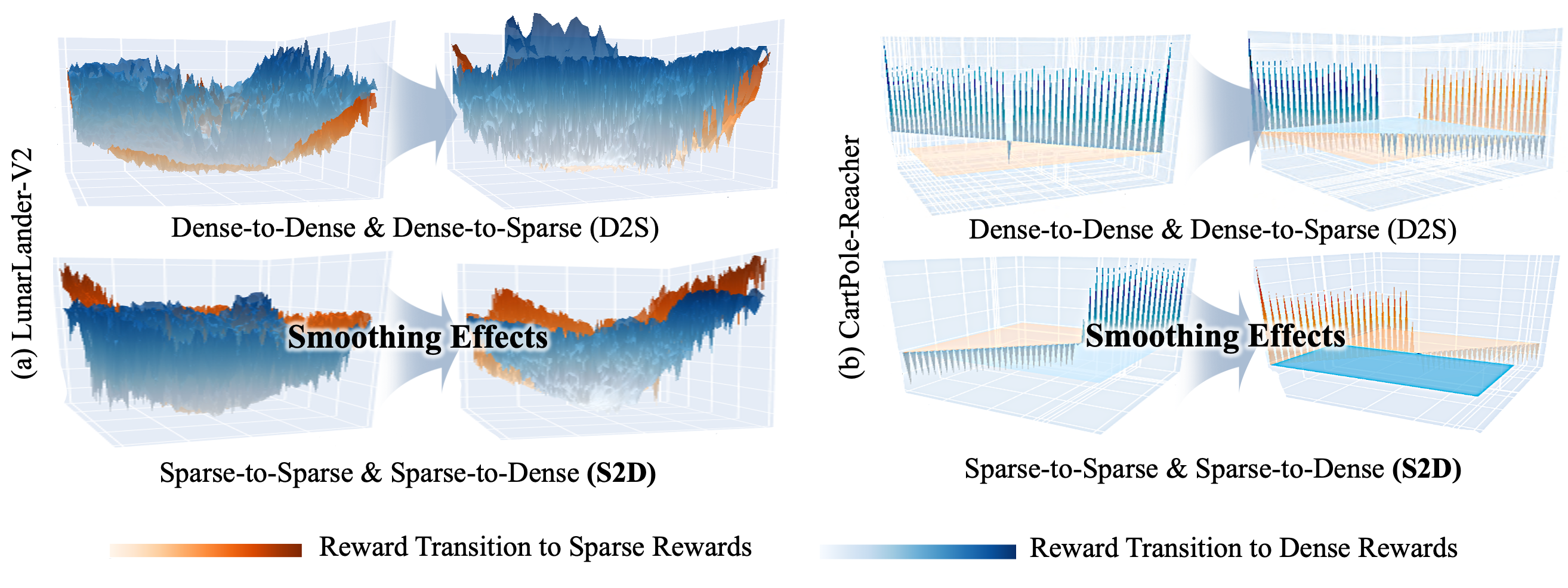}
\caption{ Analysis of policy loss landscape after reward transition. The 3D visualization depicts the policy loss landscape following a reward transition, starting with either a sparse or dense reward. 
}
\label{LossLandscape}
\end{figure*}

\subsection{Impact on 3D Policy Loss Landscape}


Our visualizations, presented in Figure \ref{LossLandscape} and detailed in Appendix B, emphasize significant \textit{smoothing effects}, especially with the S2D transition. In Figure \ref{LossLandscape}, the upper row shows dense-to-dense (Only Dense) and D2S transitions, while the lower row displays S2D and sparse-to-sparse (Only Sparse) transitions. Significant smoothing effects were primarily observed during the S2D transition, aiding in overcoming local minima and promoting wider minima, thereby enhancing generalization. These effects became evident after the transition at T = 50 and T = 2000 in LunarLander, and at T = 3500 in Cartpole-Reacher. Detailed 3D visualizations are provided in Appendix B.

While our primary experiments focused on Soft Actor-Critic (SAC)\cite{haarnoja2018soft}, we also evaluated other algorithms, such as Proximal Policy Optimization (PPO)\cite{Schulman2017ProximalPO} and Deep Q-Network (DQN)\cite{mnih2013playing}, as detailed in Appendix C, and observed similar smoothing effects during the S2D reward transitions. Moreover, to further illustrate these smoothing effects, we experimented with these other algorithms in a gridworld environment that reveals changes in the policy loss landscape more intuitively.


\subsection{Results of Wide Minima} \label{wideminimaresult} Using sharpness metrics, we analyzed the convergence behavior at the end of training for networks guided by S2D reward transitions and compared them to baseline models. Lower sharpness values, which correspond to wider minima, were found to be associated with improved generalization. As evident from Table~\ref{table:sharpness}, only agents following the S2D reward transition converged to these wider minima, indicating superior performance in various complex environments.

\subsection{Results of Tolman’s Maze Experiment}

\begin{figure*}[t!]
     \centering
    \includegraphics[width=0.9\textwidth]{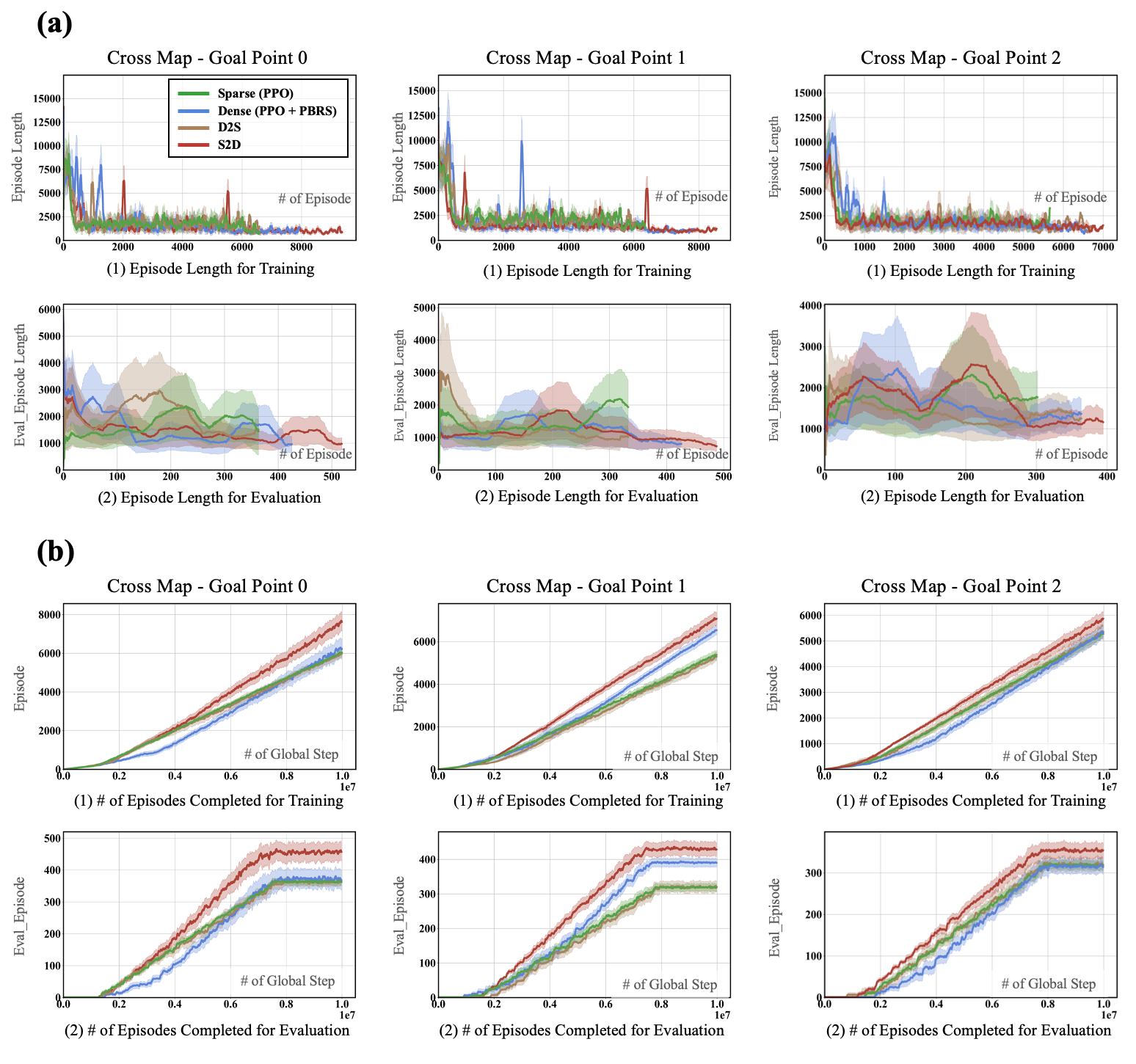}
    \caption{Performance analysis of agents using different reward strategies in the Cross maze environment. (a) Episode length during training and evaluation for Goal Points 0, 1, and 2. (b) Number of episodes completed for training and evaluation phases at different goal points. }
    \label{figure:crossmap} 
\end{figure*}

\subsubsection{Cross Maze}

We measured episode length, a performance metric used in Tolman's maze experiment. Figure~\ref{figure:crossmap}a shows that agents using the S2D reward transition achieve consistently shorter episode lengths across Goal Points 0, 1, and 2 during training and evaluation compared to other reward structures.
Consequently, the plot of S2D extended furthest along the horizontal axis, indicating that the agent completed more episodes within the same number of global steps.

To get a better understanding of learning trends, we measured the number of episodes completed during training and evaluation as a function of global steps in Figure \ref{figure:crossmap}b. All S2D agents demonstrated a steeper increase in completed episodes, even in the more challenging scenario of Goal Point 2. This indicates that S2D agents display higher sample efficiency and success rates across all scenarios, demonstrating superior performance and generalization to unseen goal positions.

\begin{figure*}[t!]
     \centering
    \includegraphics[width=0.9\textwidth]{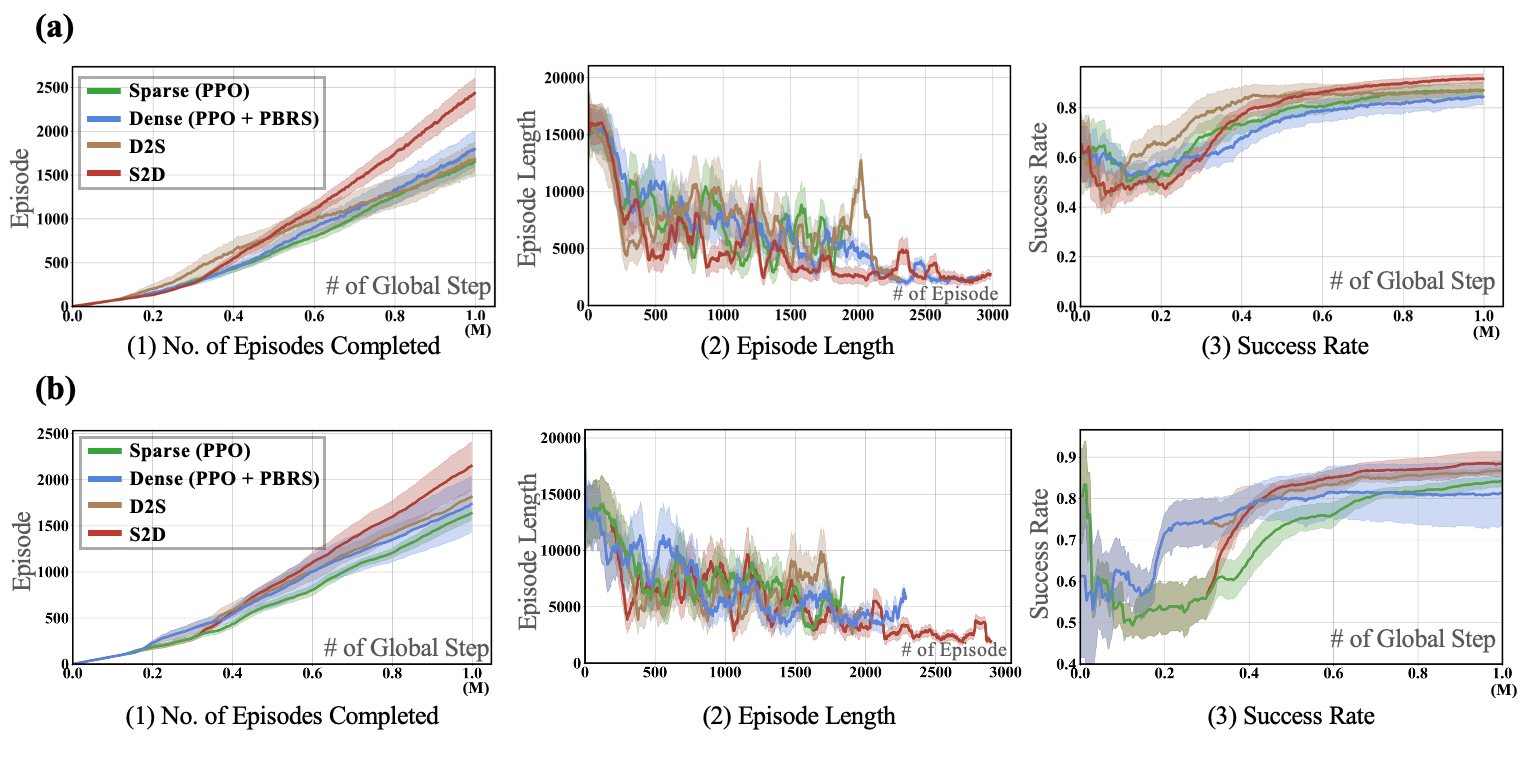}
    \caption{Performance analysis of agents using different reward strategies in the playroom maze environment. (a) Random Seed: Training results with a random seed, similar to typical experimental settings. (b) Fixed Seed: In the initial phase, before reward transitioning (at 0.3 global steps), sparse or dense rewards were used, and seeds were fixed to ensure fairness and clarity in the analysis. A total of 6 seeds were used for each experiment.}
    \label{figure:toddlerresult} 
\end{figure*}

\subsubsection{playroom maze}

Figure~\ref{figure:toddlerresult}(a),(b)-(1),(2) show that S2D agents achieve significantly shorter episode lengths during training, indicating improved sample efficiency and enhanced performance compared to other reward strategies. This suggests that the S2D reward transition mechanism effectively guides agents to reach goals faster by balancing exploration and exploitation more efficiently and accelerating learning.
Figures~\ref{figure:toddlerresult}a,b-(3) show that overall, S2D demonstrates greater stability, lower variance, and improved learning performance compared to the only dense reward strategy and others, even in visually complex, high-dimensional playroom environments. This highlights the robustness and generalization capability of the S2D approach. Notably, after the reward transition point, compared to the purely dense reward strategy, the all-S2D approach achieves faster convergence with much more stable performance, maintaining a much lower standard deviation. This is clearly observed in the success rate results from both the random seed and fixed seed experimental environments.

\subsubsection{Visualization of Trajectory}

\begin{figure*}[h!]
     \centering
    \includegraphics[width=0.9\textwidth]{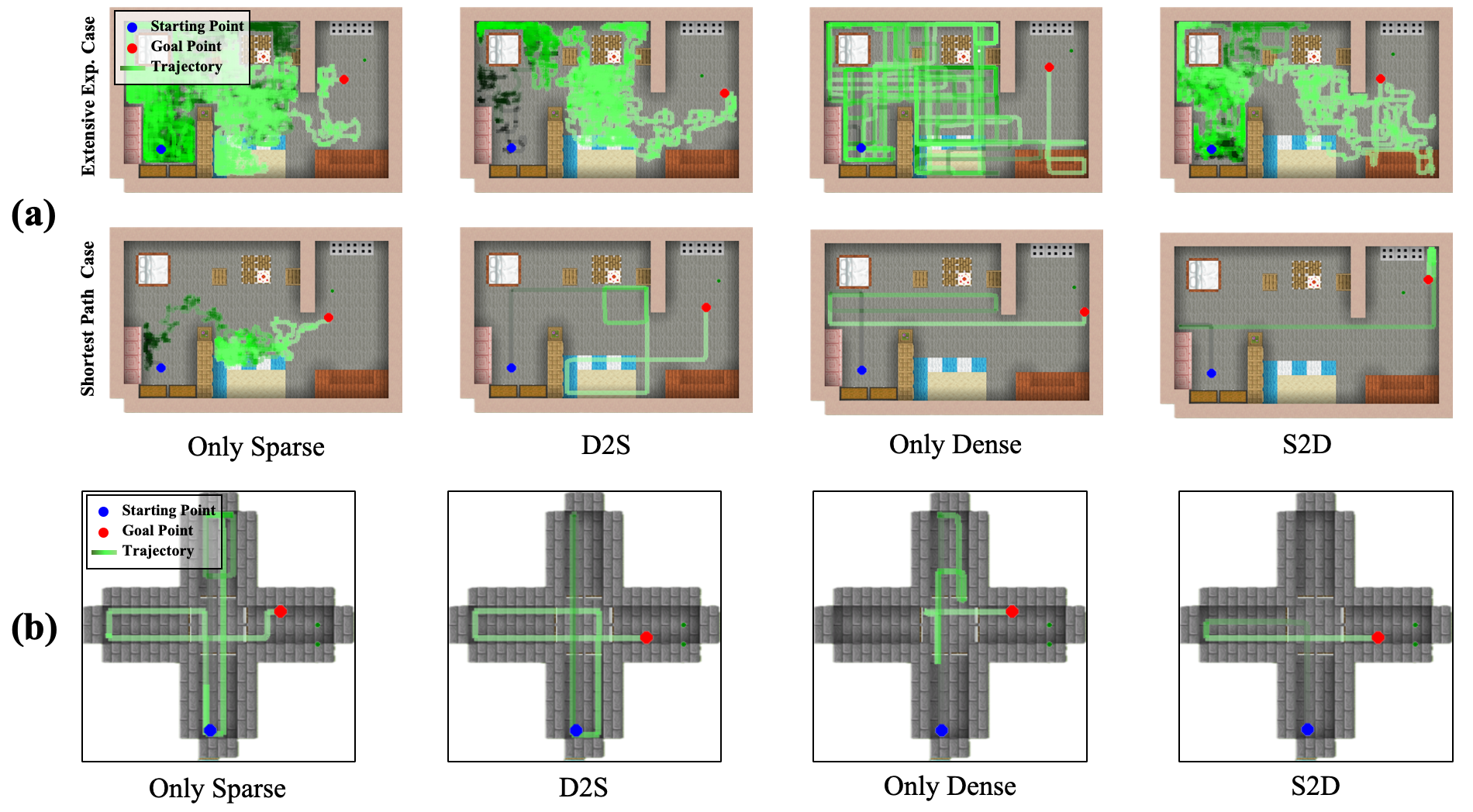}
    \caption{ Visualizations of the trajectories near the final episode and feature analysis in maze environments. (a) playroom maze Trajectories: The top row displays the exploration paths of agents with different reward settings. The bottom row illustrates the most frequent shortest paths. (b) Cross Maze Trajectories: The most frequent shortest paths are displayed. }
    \label{figure:trajectory} 
\end{figure*}

In the playroom maze (Figure\ref{figure:trajectory}-(a)), agent trajectories under different reward settings reveal significant differences in exploration behavior. The top row showcases agents’ extensive exploratory paths. S2D and Only Sparse agents exhibit diverse, exploratory trajectories, providing opportunities to robustly learn about the environment and objects from various angles. This exploration suggests that these agents can learn more about their environment, similar to how toddlers learn through extensive exploration. In contrast, Only Dense agents show more direct and angular trajectories, indicating limited exploration and a focus on reaching the goal quickly. This pattern suggests that dense reward agents focus on quickly reaching the goal, which may limit their ability to learn about the environment comprehensively. The bottom row illustrates the most frequent shortest trajectories. S2D agents show the most efficient paths to the goal, effectively balancing the exploration-exploitation trade-off. In the Cross Maze (Figure\ref{figure:trajectory}-(b)), similar patterns are observed. Agents using the S2D reward transition demonstrate better shortest trajectories.

\begin{figure*}[h!]
     \centering
    \includegraphics[width=0.9\textwidth]{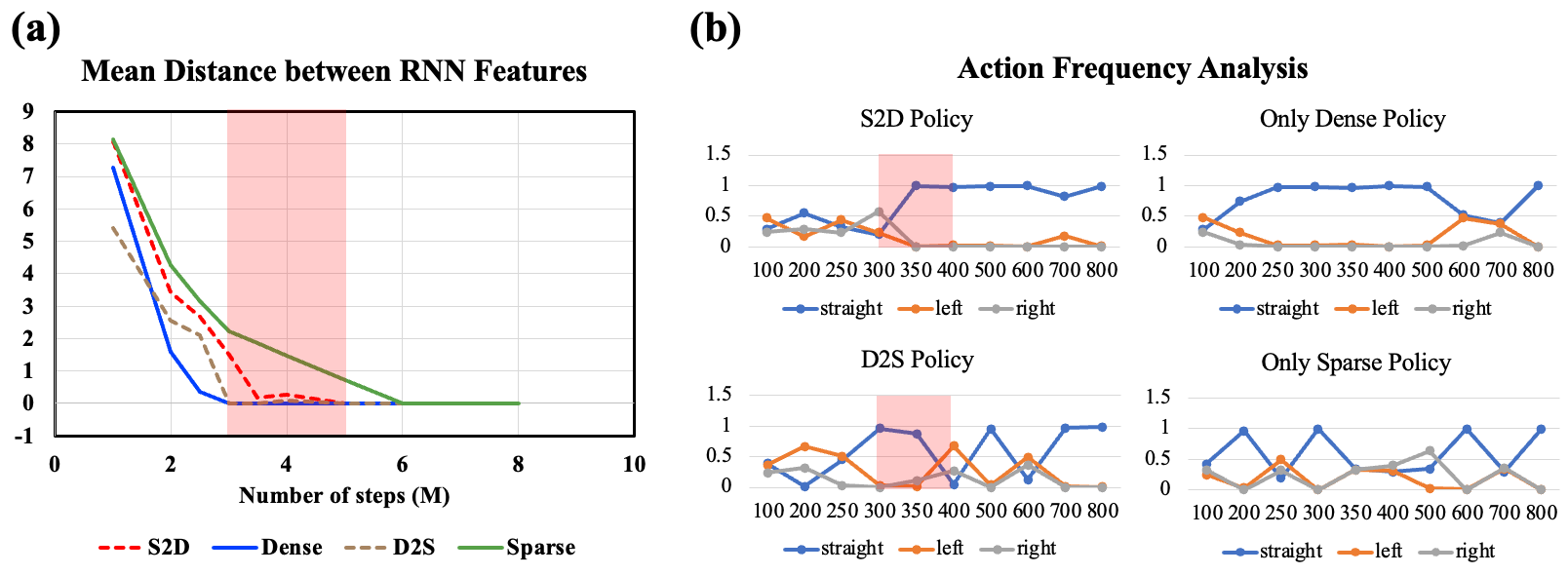}
    \caption{Feature analysis in maze environments: RNN Feature and Action Frequency Analysis. (a) The left graph shows the mean distance between RNN features during training, with the reward transition occurring at 3M steps. In the region highlighted in red, the features converge notably faster for S2D compared to Only Dense, suggesting that learning with sparse rewards initially provides good initial parameter points. (b) The right plots depict action distributions (straight, left, right). The reward transition occurred at 3M steps, and the plots are based on results from over five trials.
    }
    \label{figure:analysis} 
\end{figure*}

\subsubsection{Mean Distance Between RNN Features}

To evaluate the impact of reward transitions on RL agents' internal representations, we analyzed the convergence of RNN feature representations in the playroom maze. Figure~\ref{figure:analysis}-(a) depicts the mean Euclidean distance between hidden state vectors for agents trained with S2D, Only Dense, and Only Sparse reward settings. Agents trained using the S2D framework exhibited a significant reduction in feature distance following the reward transition, indicating faster convergence of internal representations. This suggests that the sparse reward phase serves as a foundational learning stage, fostering robust initial parameter configurations through extensive exploration and facilitating the discovery of diverse state-action mappings. These robust initial parameters enable stable and generalizable optimization during subsequent dense reward learning. In contrast, agents trained exclusively with dense rewards exhibited slower and less consistent convergence compared to those using the S2D approach. This disparity is likely due to limited exploration, which prematurely reinforces suboptimal behaviors. Agents trained solely with sparse rewards demonstrated the slowest convergence overall, as the scarcity of reward signals impeded the development of meaningful representations. For D2S agents, most convergence occurred during the dense reward phase. The sparse reward phase had minimal impact post-transition, as initial dense reward optimization induced a primary dense reward bias, thereby limiting adaptability to sparse rewards in later stages.


\subsubsection{Action Frequency Analysis}

Figure~\ref{figure:analysis}-(b) illustrates the behavior distributions of agents trained under various reward baseline models. Each colored line—blue (straight), orange (left), and gray (right)—represents the proportion of behaviors observed at specific checkpoints. During the sparse reward phase, both the S2D and Only Sparse models exhibited significant instability in policy behaviors. Upon transitioning to dense rewards, the S2D and Only Dense models displayed markedly divergent outcomes. The Only Dense model continued to show instability even after apparent convergence, suggesting that its policy may have settled into a suboptimal local minimum. This persistent instability indicates a vulnerability to environmental changes, thereby limiting the model's generalizability. In contrast, the S2D approach maintained consistent stability across all five trials, implying that its policy occupies a broader and more optimal solution space. These findings highlight the robustness of the S2D framework in developing stable and generalizable policies capable of adapting to environmental variations.




\section{Discussion}

Throughout this study, we focus on the key challenge of balancing exploration and exploitation in goal-oriented RL, particularly with reward shaping. This challenge is heightened in scenarios involving high-dimensional raw input, such as egocentric real-world environments. To address this, we explore the significant advantages of incorporating S2D reward transitions, ranging from simple gridworld environments to complex 3D egocentric-view settings, inspired by toddler learning patterns.

\subsection{Performance Improvement}
Our results consistently show that S2D outperforms other reward-shaping strategies across both discrete and continuous action spaces. In more generalizable environments like ViZDoom and mazes, S2D agents still converged faster, achieved optimal performance, and exhibited lower variance compared to reward baselines. Moreover, we observed that agents equipped with intrinsic motivation algorithms excel at discovering diverse states but mainly struggle to focus on specific goals, a critical requirement in goal-oriented RL. In contrast, the S2D transition mechanism effectively balances exploration with exploitation, thereby facilitating stronger goal attainment. Ablation studies reveal that the most beneficial point for transitioning from sparse to dense rewards typically lies around the first quarter of the early training schedule, although the precise timing depends on task complexity. For instance, UR5-Reacher requires an extended free exploration phase before transitioning, aligning with early critical learning periods observed in infant development.

\subsection{Impact on 3D Policy Loss Landscape}
One of the most striking findings of this study is the impact of S2D transitions on policy loss landscapes. Using our Cross-Density Visualizer (Figure~\ref{LossLandscape}), we observed significant smoothing effects during S2D transitions, particularly in environments requiring generalization. These effects reduce the sharp peaks and valleys typically associated with dense reward settings, thereby facilitating convergence to wider minima. While our primary experiments utilize SAC, we extended our analysis to include other algorithms, such as PPO~\cite{Schulman2017ProximalPO} and DQN~\cite{mnih2013playing}, to ensure a broader evaluation. Notably, this smoothing effect predominantly appears with the S2D transition, as further confirmed in additional gridworld experiments detailed in Appendix C.

\subsection{Link Between Wide Minima and Toddler-Inspired Reward Transition}
Wide minima, by virtue of their broad and flat characteristics, tend to produce solutions that generalize well to previously unseen environments. The sharpness metrics in Table \ref{table:sharpness} support this claim, showing that S2D agents consistently achieve lower sharpness values—indicative of wider minima. Indeed, only the S2D reward transition allowed agents to converge to the broadest minima in LunarLander and CartPole-Reacher, where even the Only Sparse approach demonstrated some success. A notable exception arises in UR5-Reacher, where the Only Sparse setting exhibits unexpectedly low sharpness but simultaneously yields near-zero performance. This outcome is likely due to limited or absent gradient updates, causing gradient stagnation and high variance—factors that can artificially reduce sharpness metrics. Nonetheless, the most critical comparison lies with the Only Dense baseline: S2D not only outperforms dense rewards but also maintains high performance while exhibiting lower sharpness, aligning it more closely with wide minima that facilitate robust generalization.! "

\subsection{Key Insights from Reinterpretation of Tolman’s Maze}

Inspired by Tolman’s maze, our investigation centers on how early free exploration under sparse reward influences policy development within the S2D framework. To this end, we designed two distinct maze environments to systematically evaluate these effects. Our experiments reveal that, across all maze scenarios, S2D agents achieve shorter episode lengths and greater sample efficiency compared to other reward configurations, such as using only dense rewards or D2S. In particular, trajectory visualizations in the Playroom Maze (Figure~\ref{figure:trajectory}) demonstrate that S2D agents exhibit more efficient behaviors, reliably identifying optimal paths compared to other approaches. 

During the sparse reward phase, S2D agents explored a broader range of pathways, whereas agents trained with only dense rewards followed more constrained and angular trajectories. This suggests that initiating training with sparse rewards, rather than relying on dense rewards from the outset, allows for more diverse experience gathering—ultimately laying a foundation for efficient policy refinement once denser rewards are introduced.

To further assess how reward transitions influence the agents’ internal representations, we measured the mean distance between RNN features during training (Figure~\ref{figure:analysis}-a). We observed that S2D agents showed a notable reduction in feature distances following the reward transition, suggesting a faster convergence of internal representations compared to the Only Dense group. In contrast, Only Dense agents—without the benefits of initial free exploration—experienced slower and less consistent convergence. Furthermore, policy visualization (Figure~\ref{figure:analysis}-b) reinforces this observation, highlighting the stable exploration-exploitation balance maintained by S2D agents.

Concluding these findings from Section 6.4.4, we claim that the introduction of sparse rewards at the outset promotes the development of robust initial parameter settings. These parameters accelerate stable and generalizable policy learning when dense rewards are introduced later. This approach aligns with Tolman’s original hypothesis: initial free exploration, followed by the introduction of stronger stimuli, such as rewards, leads to optimal performance outcomes.
 

\section{Conclusion}
Drawing inspiration from developmental learning of toddlers, this research advances a dynamic reward transition model in goal-oriented RL, challenging the traditional static reward densities. Transitioning from S2D rewards improves learning efficiency across various RL tasks while also fundamentally changing how the agent learns, encouraging a smoother, more stable progression toward optimal behaviors. Our Cross-Density Visualizer reveals a key smoothing effect on the policy loss landscape during these transitions, and sharpness metrics confirm that S2D fosters wider minima, promoting better generalization. Further, our reinterpretation of Tolman’s maze experiments within custom 3D egocentric environments underscores the critical role of early free exploration in establishing good initial policy parameters—akin to a cognitive map—which optimizes subsequent navigation as dense rewards are introduced. This integration of developmental insights into RL methodology paves the way for designing more adaptable, high-performance learning systems, significantly contributing to the field of RL.

\section{Future Work and Opportunities}

The integration of the toddler-inspired reward transition paradigm within reinforcement learning (RL) frameworks has established a foundational groundwork, demonstrating that the Sparse-to-Dense (S2D) transition can enhance agent generalization and performance. Building upon this foundation, several promising avenues remain for further exploration, presenting significant opportunities to advance the S2D framework and its applications.

\subsection{Automating Reward Transition Timing} A pivotal area for future research is the development of automated methods to determine the optimal timing for reward transitions. Currently, the transition from sparse to dense rewards is manually scheduled based on predefined criteria. Our preliminary investigations into smoothing effects and the convergence of recurrent neural network (RNN) representations lay the groundwork for automated optimization methods. Future work should explore adaptive scheduling algorithms or meta-learning approaches that dynamically adjust the reward transition timing based on real-time assessments of policy loss landscapes and representation convergence metrics. Automating this process would enhance the adaptability and efficiency of the S2D framework, reducing the reliance on manual intervention and enabling more nuanced reward shaping tailored to the agent's learning progression.

\subsection{Integrating with Model-Based RL Frameworks} Another promising direction involves the integration of the S2D reward transition with model-based reinforcement learning (RL) approaches. Model-based RL, which leverages internal representations of the environment to predict future states and outcomes, contrasts with model-free RL, where agents learn policies directly from interactions without explicit environmental models. While our Tolman maze experiments utilized model-free RL settings, incorporating model-based methods could enable more informed decision-making by utilizing these predictive models. By combining S2D transitions with model-based frameworks, future research can directly analyze and compare the impact of reward transitions on representation learning. This integration could facilitate the development of more human-like learning environments, where agents not only learn from rewards but also build predictive models of their surroundings, enhancing both efficiency and adaptability.

\subsection{Extending to Multi-Agent Systems and Real-World Applications} Expanding the S2D framework to multi-agent systems and real-world applications represents another significant opportunity for future research. In collaborative tasks, where agents must balance individual goals with group objectives, dynamic reward transitions could foster effective cooperation and healthy competition. Additionally, applying the S2D framework to real-world scenarios, such as robot AI, would allow for the validation and refinement of our approach in more practical and complex environments. This extension could lead to the development of more sophisticated and robust RL frameworks capable of handling the intricacies of real-world interactions and multi-agent dynamics.

\section{Acknowledgments}
The authors would like to express their sincere gratitude to Inwoo Hwang, Changhoon Jeong, Moonhoen Lee, and Dong-Sig Han for their insightful discussions and valuable suggestions on the early drafts of this paper. This work was partly supported by the IITP (2021-0-02068-AIHub/15\%, 2021-0-01343-GSAI/10\%, 2022-0-00951-LBA/15\%, 2022-0-00953-PICA/25\%) and NRF (RS-2023-00274280/10\%, 2021R1A2C1010970/25\%) grant funded by the Korean government.

\bibliographystyle{plainnat}
\bibliography{aaai24}


\onecolumn
\section*{Supplementary Material: Insights into Toddler-Inspired Reward Transitions in Goal-Oriented Reinforcement Learning}
\setcounter{page}{1}

\begin{mdframed}[frametitlealignment=\center,leftline=false, rightline=false]
\begin{itemize}[leftmargin=*]
\item \textbf{Part A:} This part elaborates on the experimental setups and additional appendices referenced in the main paper.

\item \textbf{Part B:} This part showcases detailed results of the 3D policy loss landscape visualizations post-stage transition for Toddler-inspired S2D Reward Transition, in comparison to various baselines. This complements the section: Visualizing Post-Transition 3D Policy Loss Landscape: Cross-Density Visualizer in the main text.

\item \textbf{Part C:} This part includes extra experiments, analyses, and further visualizations of the 3D policy loss landscape across different algorithms in a gridworld setting.
\end{itemize}
\end{mdframed}

\section{Section A: Experimental Details}
\subsection{Comparison of Overall Experimental Setup}

Table 3 summarizes the experimental environments used in our study. Environments above the double line are discussed in the main text, while those below are detailed in the appendices. Each setup was tailored to the Toddler-inspired S2D reward transition, with specifics provided in the environment setup sections.
\begin{table}[h]
    \renewcommand*{\arraystretch}{1.1}
    \begin{center}
        \caption{This table compares the experimental environments utilized in our research. The environments above the double line are covered in the main body, while those below are included in the appendices. Each environment and reward scheme was customized to align with the Toddler-inspired S2D reward transition, with full details provided in the respective environment setup sections.
.}
        \label{table:detailenv}
        \begin{adjustbox}{width=\textwidth}
            \begin{tabular}{c|cccc}
            \Xhline{6\arrayrulewidth}
            \multirow{2}{*}{Environment}& Task  & Difficulty Settings &Environments Type & Input\\
            & \# of Stages & Point of View & Action Space & Observation Types\\
            \Xhline{6\arrayrulewidth}
            \rowcolor[HTML]{EEEEEE}
            &LunarLander-V2&-&2D& Coordinate \& Velocity \& Angle \& Boolean flag value \\
            \rowcolor[HTML]{EEEEEE}
            \multirow{-2}{*}{\cellcolor[HTML]{EEEEEE} OpenAI Gym\citep{todorov2012mujoco}}&2-stage  & StaticView  &Continous& State-based RL\\
            \multirow{2}{*}{MuJoCo\citep{todorov2012mujoco}} &CartPole-Reacher&-&3D& Joint Value \& Goal Position\\
            &2-stage &StaticView&Continuous& State-based RL\\
            \rowcolor[HTML]{EEEEEE}
            &UR5-Reacher&-&3D& Joint Value \& Goal Position\\
            \rowcolor[HTML]{EEEEEE}
            \multirow{-2}{*}{\cellcolor[HTML]{EEEEEE} MuJoCo\citep{todorov2012mujoco}}&2-stage &StaticView&Continuous& State-based RL\\
            &  Seen \& Unseen Navigation & - &3D& RGB-D\\
            \multirow{-2}{*}{ViZDoom\citep{kempka2016vizdoom}} & 2-stage &Egocentric View&Discrete& Visual RL\\
           \rowcolor[HTML]{EEEEEE}&Toddler playroom \& Cross Maze &-&3D& RGB\\
            \rowcolor[HTML]{EEEEEE}
            \multirow{-2}{*}{Minecraft}& 2-stage& Egocentric View& Discrete&Visual RL\\
            \hline
            \hline
            \rowcolor[HTML]{EEEEEE}
            &Shelf-delivery&Level3&2D& Internal state of the surrounding tiles\\
            \rowcolor[HTML]{EEEEEE}
            \multirow{-2}{*}{RWARE\citep{papoudakis2021benchmarking}}&Non-humanoid&StaticView& Discrete&3-stage \& State-based RL\\
            \rowcolor[HTML]{FFFFFF}
            &  Navigation & - &2D&  Position-based value\\
            \rowcolor[HTML]{FFFFFF}
             \multirow{-2}{*}{\cellcolor[HTML]{FFFFFF} Gridworld} &2-stage &StaticView&Discrete& State-based RL\\
            \Xhline{6\arrayrulewidth}
            \end{tabular}
        \end{adjustbox}
    \end{center}
\end{table}

\begin{table}[h]
    \centering
    \renewcommand{\arraystretch}{1.1}
    \caption{Detailed settings for hyperparameter $N$, indicating the number of frames after which the stage transition occurs for each environment ($\mathscr{C}_1$, $\mathscr{C}_2$, and $\mathscr{C}_3$). In ViZDoom experiments, $N$ represents the number of updates, while for Gridworld-DQN, $\mathscr{C}_1$=100, $\mathscr{C}_2$=200, and $\mathscr{C}_3$=300 episodes.}
    \begin{adjustbox}{width=0.7\textwidth}
        \begin{tabular}{c|cccc}
        \hline
        Environment & Total \# of Training & 1$N$($\mathscr{C}_1$) & 2$N$($\mathscr{C}_2$) & 3$N$($\mathscr{C}_3$) \\
        \hline
        \rowcolor[HTML]{EEEEEE} 
        LunarLander-V2 & 1M frames & 100k & 200k & 400k \\
        ViZDoom-Seen \& Unseen & 1M frames & 50k & 100k & 250k \\
        \rowcolor[HTML]{EEEEEE} 
        CartPole-Reacher & 12k episodes & 1k & 2k & 3k \\
        UR5-Reacher & 25k episodes & 1k & 2k & 3k \\
        \rowcolor[HTML]{EEEEEE} 
        Toddler Playroom Maze &  10M frames & 1M & 2M & 3M \\
        Cross Maze & 10M frames & 1M & 2M & 3M \\
        \hline
        \hline
        \rowcolor[HTML]{EEEEEE} 
        RWARE & 7M frames & 1M & 2M & 3M \\
        \rowcolor[HTML]{FFFFFF} 
        Gridworld & 25k episodes & 3k & 5k & 7k\\
        \hline
        \end{tabular}
    \end{adjustbox}
    \label{table:hyperparams}
\end{table}

\subsection{Reward Transition Hyperparameters}
\begin{figure}[thb]
\centering
    \includegraphics[width=1\columnwidth]{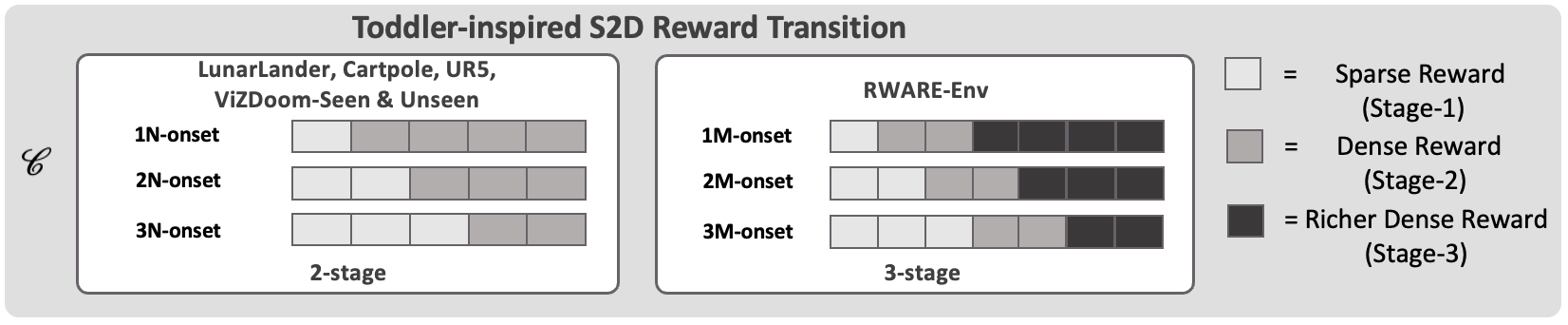}
    \caption{Visualization of the overall setup, including the number of stages and the transition times, in Toddler-inspired S2D experiments across all environments.}
   \label{figure:Toddler-inspired S2D}
\end{figure}

In Table~\ref{table:hyperparams} and Figure~\ref{figure:Toddler-inspired S2D}, we pinpoint the exact moment when the agent transitions to dense reward stages across various environments, along with the total number of stages for each.

\subsection{Model Hyperparameters}
\label{app:hyper}
The Table~\ref{table:hyper3} provides information on the hyperparameters for each environment used in our study.

\begin{table}[h]
    \renewcommand*{\arraystretch}{1.1}
    \centering
    \caption{The hyperparameters for our experiments and those mentioned in the Appendices are provided here. When visualizing the policy loss landscape for LunarLander, we utilized discount factors $\gamma$ of 1 and 0.99.}
    \begin{adjustbox}{width=\textwidth}
        \begin{tabular}{c|ccccccccc}
        \Xhline{3\arrayrulewidth}
        Hyperparameters & LunarLander & CartPole-Reacher & UR5-Reacher & ViZDoom-S\&U & Minecraft & RWARE & Gridworld  \\
        \Xhline{3\arrayrulewidth}
        \rowcolor[HTML]{EEEEEE} 
        RL algorithms &  SAC & SAC & SAC & A3C & PPO & PPO & PPO  & \\
        Learning Rate & 3e-4 & 0.0007 & 0.0007 & 7e-5 & 3e-4 &5e-4 & 5e-4  \\ 
        \rowcolor[HTML]{EEEEEE} 
        Value Function Coefficient & 3e-4 & - & - & 0.5 & 0.5 & - & 5e-4 & \\
        Discount Factor & 0.99 & 0.99 & 0.99 & 1.0 & 0.99 &0.99 & 0.99 & \\
        \rowcolor[HTML]{EEEEEE} 
        Batch Size & 128 & 128 & 128 & - & 128 & 10 & 128 \\
        Optimizer & Adam & Adam & Adam & Adam & Adam & Adam & Adam \\
        \rowcolor[HTML]{EEEEEE} 
        Maximum \# of Steps & 500 & 200 & 500 & 50 & 20000 & 150 & 50 \\
        Entropy Coefficient & 0.2 & Auto & Auto & 0.1 & 0.005 &0.01 & 0.03 \\
        
        \Xhline{3\arrayrulewidth}
        \end{tabular}
    \end{adjustbox}
    \label{table:hyper3}
\end{table}



\subsection{Environment Details}
For each environment, both the baseline methods and the proposed approach were executed using at least over five random seeds. The hardware setup included four NVIDIA GeForce RTX 3090 GPUs, two NVIDIA GeForce RTX 2080ti GPUs, and an AMD Ryzen Threadripper 3960X 24-core processor. Additionally, a total of 188 GB of RAM was utilized.

\subsubsection{OpenAI Gym: LunarLander-V2.}
In this scenario, the lander starts mid-air with random speed and orientation. The agent's main task is to control the engines to land between two flags. The optimal landing should be centered on the pad, as vertical as possible, and at low speed. Rewards are given for actions like descending from the top of the screen, achieving a gentle landing with low speed, and making contact with each leg. Penalties are applied for excessive use of the main engine to encourage fuel efficiency, and severe penalties are given for crashes or landings far from the target pad. In our experiments, we utilized the environment's default rewards as the sparse reward structure and incorporated distance-based, potential-driven dense rewards as part of our Toddler-inspired S2D reward transition, as outlined earlier.

Our reward shaping approach employs a distance-based potential function. Unlike other environments we used, the LunarLander scenario offers rewards for landing anywhere on the surface. Thus, when the lander approaches within 0.3 units of the ground (where 1.0 is the total screen height), as shown in Table~\ref{table:hyper3}, an additional potential-based reward is granted per frame.



\subsubsection{MuJoCo: CartPole-Reacher \& UR5-Reacher tasks.}
\label{mmp}

\begin{figure}[h!]
    \begin{center}
    \includegraphics[width=0.8\textwidth]{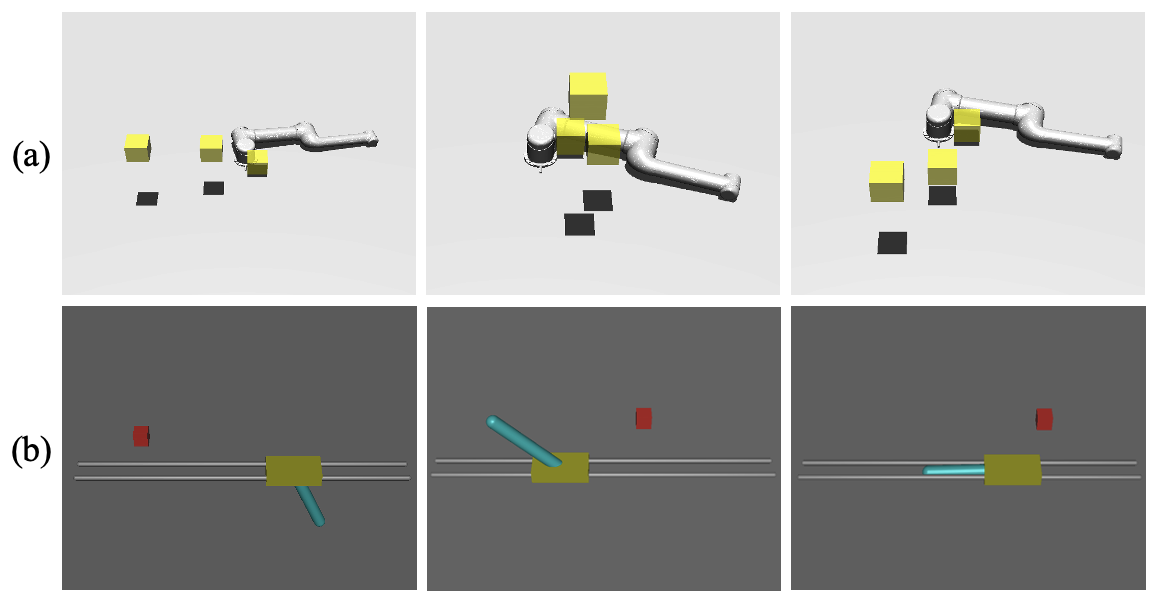}
    \caption{Examples of environments where goals randomly spawn. (a) UR5-Reacher. (b) CartPole-Reacher. }
    \label{ur5cart}
    \end{center}
\end{figure}

\begin{figure}[!htb]
    \begin{center}
    \includegraphics[width=0.8\textwidth]{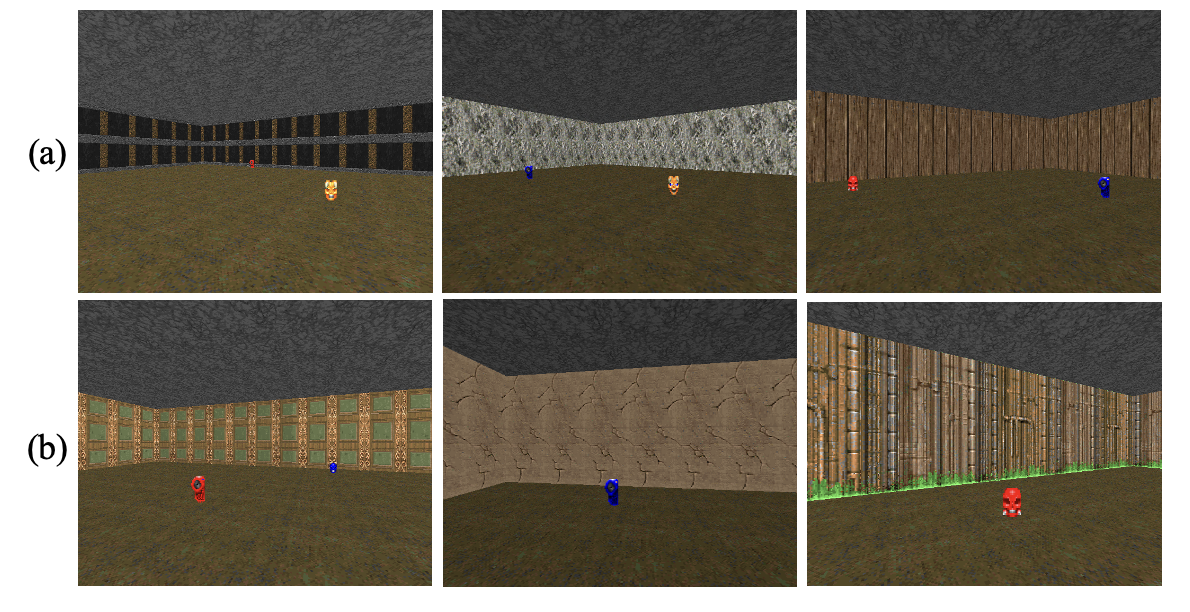}
    \caption{Egocentric views of a ViZDoom agent in environments with various walls and objects. (a) Three walls in ViZDoom-Seen. (b) Three walls in ViZDoom-Unseen.}
    \label{fig:viz-seen}
    \end{center}
\end{figure}

To determine how well the Toddler-inspired S2D reward transition works across different tasks, we leveraged the MuJoCo \cite{todorov2012mujoco} engine, a popular physics simulator for virtual environments. Our study focused on demanding continuous control tasks with goals that change randomly each episode, requiring the agent to adapt and reach the target.

In the CartPole-Reacher task, the agent controls a cart moving along a horizontal line to keep an attached pole balanced. Since the default task is relatively easy for agents to master, we raised the difficulty by setting a goal that demands the pole's end to be within a specific radius. This goal is placed on the upper side of the horizontal line, making it a challenging yet achievable target, as depicted in Figure~\ref{ur5cart}-(b).

The UR5-Reacher task involves a robotic arm with six degrees of freedom, with each joint allowing movement along one axis. This setup provides the arm with extensive flexibility to reach various positions and orientations, but also presents a complex control challenge. In this task, the agent must learn to maneuver the arm to reach a specific location, with the goal randomly assigned in each episode, as shown in Figure~\ref{ur5cart}-(a).

For both UR5-Reacher and CartPole-Reacher tasks, we implemented a potential-based dense reward system from the beginning, based on the distance between the agent's current state and the target state, as illustrated in Table~\ref{ur5cart}. To encourage faster task completion, we also applied a living penalty.

\subsubsection{ViZDoom-Seen \& Unseen.}
ViZDoom \cite{kempka2016vizdoom}, a simulator derived from the first-person shooter game Doom, was developed to advance research in RL. In this study, we utilize and adapt the egocentric navigation task from \cite{kim2021goal}. The task requires the agent to begin in a corner of a square room and navigate to the correct object out of two present in the room. The target object is randomly selected at the start of each episode and is known to the agent.

The two objects, \textit{Card} and \textit{Skull}, each come in three different colors (Red, Blue, Yellow) to prevent the agent from memorizing based solely on color. Additionally, the map features three distinct wall textures in both Seen and Unseen variations, with unique textures for each version.

The agent's input consists of a series of four RGB-D frames, each with a resolution of 42x42 pixels. The agent can select from three discrete actions: turning clockwise, turning counterclockwise, and moving forward. Each action is repeated over four in-game frames, and in our manuscript, a single "step" is defined as these four consecutive frames.

\begin{figure}[!htb]
    \begin{center}
    \includegraphics[width=0.9\textwidth]{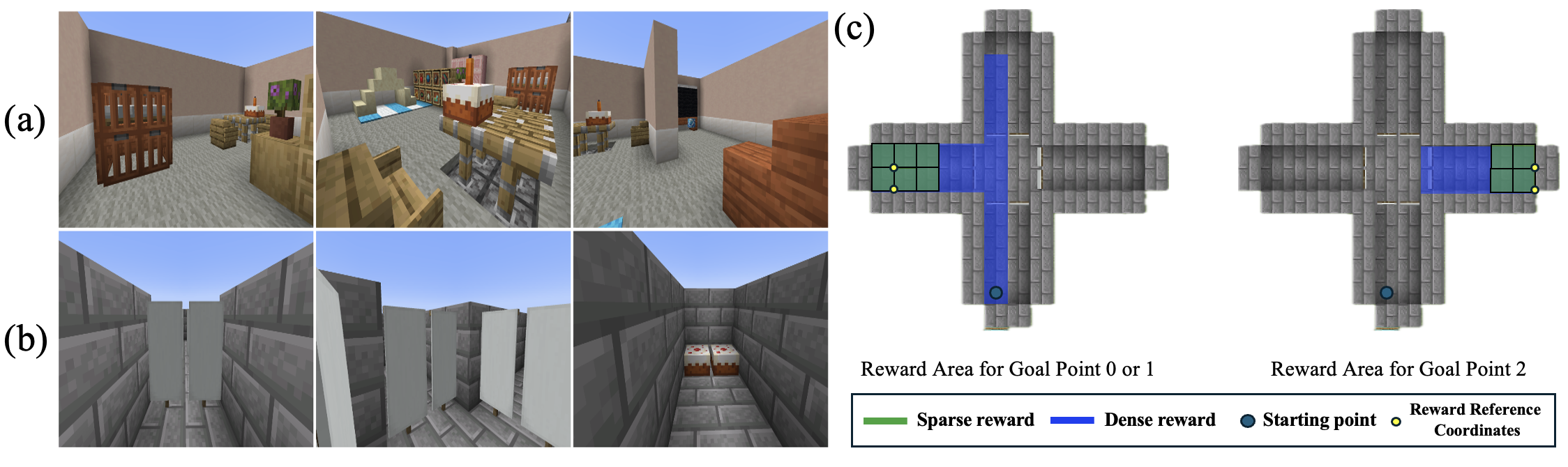}
    \caption{Egocentric views of a Minecraft agent in environments with various walls and objects. (a) Scenes in Toddler-Playroom. (b) Scenes in Cross-Maze. (c) Reward Areas for Goal Points in a Maze Environment.** The left panel shows the reward area for Goal Points 0 or 1, with dense rewards (blue) extending from the starting point and sparse rewards (green) at the goal, offering a straightforward path for exploration. The right panel highlights the reward area for Goal Point 2, which presents a challenge due to its left-skewed position and reduced reward area, requiring precise navigation for optimal reward collection.}
    \label{fig:tolmanenv}
    \end{center}
\end{figure}
In terms of rewards, the agent earns 10 points for reaching the target object and loses 1 point for selecting the wrong object. Contact with either object ends the episode, and the episode will otherwise conclude after 50 steps with a penalty of -0.1. To accelerate training, a small penalty of -0.01 is applied at each time step. To introduce visual complexity and assess the generalization capabilities of the trained agent, we use \textbf{Seen} and \textbf{Unseen} map versions, which differ in wall textures, as illustrated in Figure~\ref{fig:viz-seen}. Additionally, a dense reward of 5.0 is awarded when the agent approaches within 100 units of the goal object, given that the map measures 700 by 700 units.

We used the A3C algorithm \cite{mnih2016asynchronous} and the architecture from \cite{kim2021goal}\footnote{https://github.com/kibeomKim/GACE-GDAN} and \cite{kim2023sa,kim2023visual} as our baseline. All ViZDoom experiments were performed on two separate hardware configurations, with additional experiments on a unified hardware setup to follow.

\subsubsection{Minecraft-Toddler Playroom Maze \& Cross Maze}
We utilized an environment based on Minecraft. In the Toddler Playroom Maze experiment, the goal object is randomly positioned within a predefined goal zone, requiring the agent to generalize its navigation behavior in Maze. Within the environment, the goal is marked by a light blue glazed terracotta block display object, scaled to 0.5. Additionally, the room is decorated with various colored blocks to enhance visual richness as seen in Figure \ref{fig:tolmanenv}-(a).

The agent receives a frame of RGB egocentric vision input without a HUD (Head-Up Display). Each vision channel has a width of 114 pixels and a height of 64 pixels. The agent can choose among three discrete actions as in VizDoom: turning clockwise, turning counterclockwise, and moving forward.

As a sparse reward function, the agent receives a reward of 1 when it reaches within a Manhattan distance of 2 from the goal. For a dense reward, the agent receives a reward of 0.001 when it is within a Manhattan distance of 5 from the goal and closer than the last step. The reward is withdrawn when the agent moves farther away.

The second scenario features a cross maze where agents start at the south end and move towards three goal points, labeled clockwisely as Goal 0, 1, and 2. During training, the agent moves toward two randomly selected goals, leaving the remaining one goal point for evaluation. Then, in the evaluation phase, agents are tested on all three goal points.  Two cake blocks are placed at the goal to visually indicate it. The inputs, output actions, and reward functions are the same as in the Toddler Playroom Maze experiment. To effectively test the exploration and exploitation trade-off, a curtain is placed in the center of the maze, blocking the view in four directions as shown in Figure \ref{fig:tolmanenv}-(b). This prevents the agent from immediately seeing the goal, requiring it to explore more before exploiting the information gathered to reach the goal. Specifically, Goal Point 2 is designed to be more challenging than Goal Points 0 and 1. In this scenario, the agent is spawned from the left and is given a much reduced reward area as illustrated in Figure \ref{fig:tolmanenv}-(c). As a result, the s2d agent performed well across all goal points.

\begin{figure}[!htb]
    \begin{center}
    \includegraphics[width=0.8\textwidth]{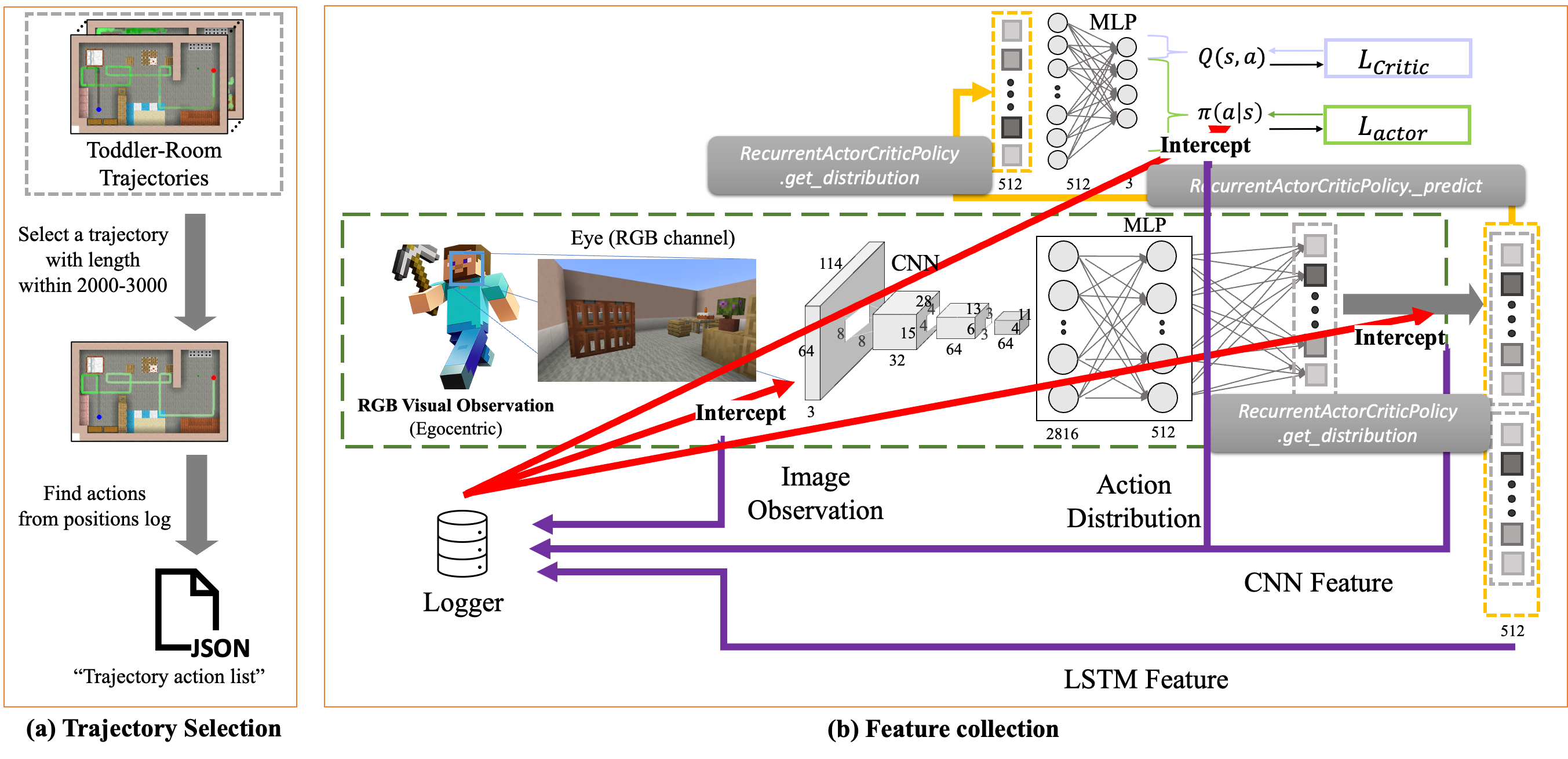}
    \caption{Data Extraction and Processing for RNN Features and Policy Visualization. (a) The process begins by selecting a trajectory where the agent reaches the goal within 2000 to 3000 steps. (b) We intercept key functions in Stable-baselines3 library to log intermediate calculation results. Observations pass through the neural network, with intermediate results logged and observations saved as image files. JSON files and image files are then generated and repeated across all model checkpoints to help visualizing the mean distance between RNN features and policy decisions.}
    \label{figprocess}
    \end{center}
\end{figure}

\begin{algorithm}
\caption{Measuring Mean Distance Between RNN Features}
\label{alg:mean_distance}
\begin{algorithmic}[1]
\REQUIRE Trajectory $\tau = \{o_1, o_2, \ldots, o_T\}$, Training intervals $I = \{i_1, i_2, \ldots, i_N\}$
\ENSURE Mean pairwise distances $D = \{d_1, d_2, \ldots, d_N\}$
\STATE Initialize empty list $D$
\FOR{each training interval $i \in I$}
    \STATE Load RNN parameters $\theta_i$
    \STATE Initialize empty list $H$
    \STATE Reset RNN hidden states
    \FOR{each timestep $t \in \tau$}
        \STATE Extract features from observation $o_t$ using CNN
        \STATE Compute hidden state $h_t = \text{RNN}(\text{features}; \theta_i)$
        \STATE Append $h_t$ to $H$
    \ENDFOR
    \STATE Compute mean pairwise distance $d_i = \text{MeanPairwiseDistance}(H)$
    \STATE Append $d_i$ to $D$
\ENDFOR
\RETURN $D$
\end{algorithmic}
\end{algorithm}

\subsubsection{Data Extraction for Mean Distance Between RNN Features and Policy Visualization}

To extract data for visualizing the mean distance between RNN features and policy decisions, we follow a structured process as seen in Figure \ref{figprocess}:

First, we select trajectories where the episode ended by reaching the goal within 2000 to 3000 steps. From these trajectories, we create an action array corresponding to the selected trajectories. Next, we hook into the \texttt{Stable-baselines3} library\cite{stable-baselines3} the following two functions to log intermediate calculation results: \texttt{RecurrentActorCriticPolicy.get\_distribution} and \texttt{RecurrentActorCriticPolicy.\_predict}.

Using the action array, we perform a rollout to move the agent. During this process, observations are passed through the agent's neural network, and intermediate calculation results are logged. Observations are also saved as image files.

Finally, the JSON files and image files generated during the rollout are used for further analysis. This process is repeated for all model checkpoint files for comprehensive visualization.





\newpage
\section{Section B: 3D Visualization of the Policy Loss Landscape After Stage Transition}
\label{3dpolicydetal}

This section provides a 3D visualization of the policy loss landscape after transitioning from initial sparse or dense reward settings to two different scenarios: one featuring sparse rewards and another with dense rewards. We employ the \textit{Cross-Density Visualizer} to map this landscape within a shared parameter space. In our visualization, the hyperparameters $\alpha$ and $\beta$ in $\tilde{\theta} = \theta + \alpha \mathbf{x} + \beta \mathbf{y}$ are set to range between -10 and 10. This setup results in two distinct datasets: Sparse-to-Dense (S2D) and Sparse-to-Sparse (Only Sparse) form one set, while Dense-to-Sparse (D2S) and Dense-to-Dense (Only Dense) form the other. We observe a noticeable smoothing effect, particularly with the Toddler-inspired S2D reward transition, which could help navigate local minima and lead to broader minima.

\subsection{Results.} Our findings show that the Toddler-inspired S2D reward transition results in a significant smoothing effect, particularly in reducing the depth of local minima. This effect is evident in the blue landscapes of Figures \ref{fig:lunar_S2D}, \ref{fig:cart_S2D}, and \ref{fig:UR5_S2D}, especially in the segments depicting sparse-to-sparse and sparse-to-dense visualizations, compared to the D2S, Only Dense, and Only Sparse methods.

The observed reduction in local minima depth suggests that agents can more readily escape local minima, leading to improved generalization performance on broader minima. To validate this hypothesis, we measured the end-of-training convergence of neural networks guided by Toddler-inspired S2D, utilizing sharpness metrics to evaluate their tendency toward wider minima compared to baseline models. As displayed in Table~\ref{table:sharpness}, agents employing the Toddler-inspired S2D reward transition demonstrate superior performance in dynamic environments by converging on broader minima.

These findings imply a direct link between the smoothing effect on the local loss landscape and the enhanced ability to escape local minima.

\subsection{Additional Insights: Visualizing Policy Loss Landscape After Reward Transition}
To gain a more profound understanding of how reward transitions affect agent behavior, we visualized the policy loss landscape following these transitions. This examination offers detailed insights into the model’s optimization landscape, highlighting specific challenges and advantages that impact continuous learning.

\subsubsection{Unique Characteristics of LunarLander-V2’s Landscape}

\begin{figure}[H]
\begin{center}
\includegraphics[width= 5\textwidth, height=0.75\textheight, keepaspectratio]{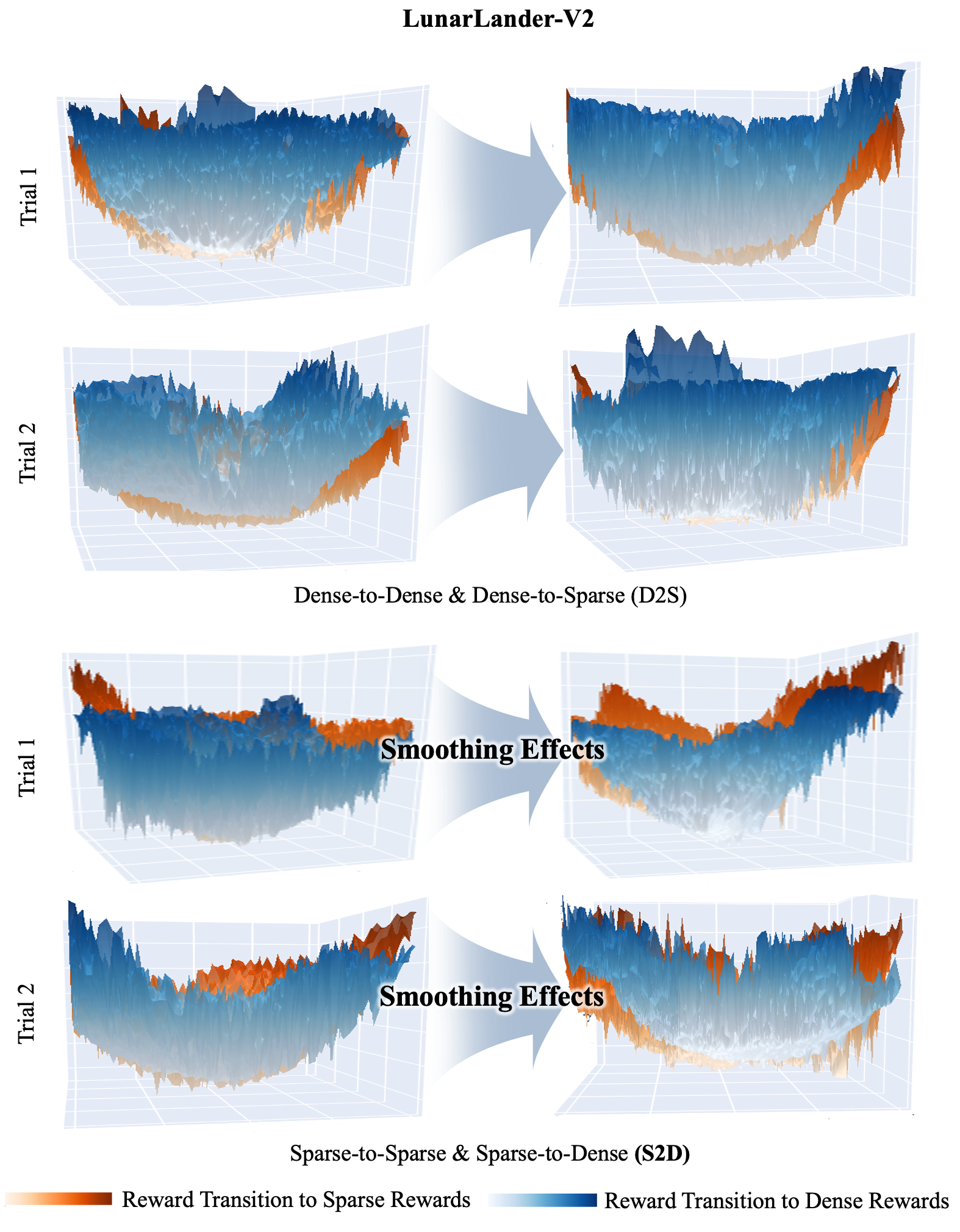}
\caption{This figure offers a detailed look at the 3D policy loss landscape during reward scheme transitions. On the left, the landscape immediately after the transition is shown, while the right side portrays the landscape around t = 2000. The initial set of lines highlights the dense-to-sparse (D2S, $\mathscr{C}_1$, red) and dense-to-dense (Only Dense, blue) transitions. Conversely, the below set focuses on sparse-to-dense (\textcolor{blue}{Toddler-inspired S2D}, $\mathscr{C}_2$, blue) and sparse-to-sparse (Only Sparse, red) transformations. Significantly, the Toddler-inspired S2D approach reveals a more pronounced reduction in the depth of local minima, indicating substantial smoothing effects across various updates.}
\label{fig:lunar_S2D}
\end{center}
\end{figure}

The LunarLander-V2 environment, depicted in Figure ~\ref{fig:lunar_S2D}, is distinguished by its unique reward distribution. Here, actions such as descending from the top of the screen to the landing pad or achieving a stable landing state yield substantial rewards ranging from 100 to 140 points. Conversely, deviations from the landing pad or crashes incur penalties. We hypothesize that this variety of reward opportunities creates a policy loss landscape for the agent that is smoother and less spiky compared to landscapes in other environments. Figure \ref{fig:lunar_S2D}: 3D Policy Loss Landscape for LunarLander-V2 showing smoothing effects in the loss landscape with the S2D reward transition.

\subsubsection{Distinct Peaks in CartPole-Reacher and UR5-Reacher}

\begin{figure}[H]
\begin{center}
\includegraphics[width=\textwidth, height=0.8\textheight, keepaspectratio]{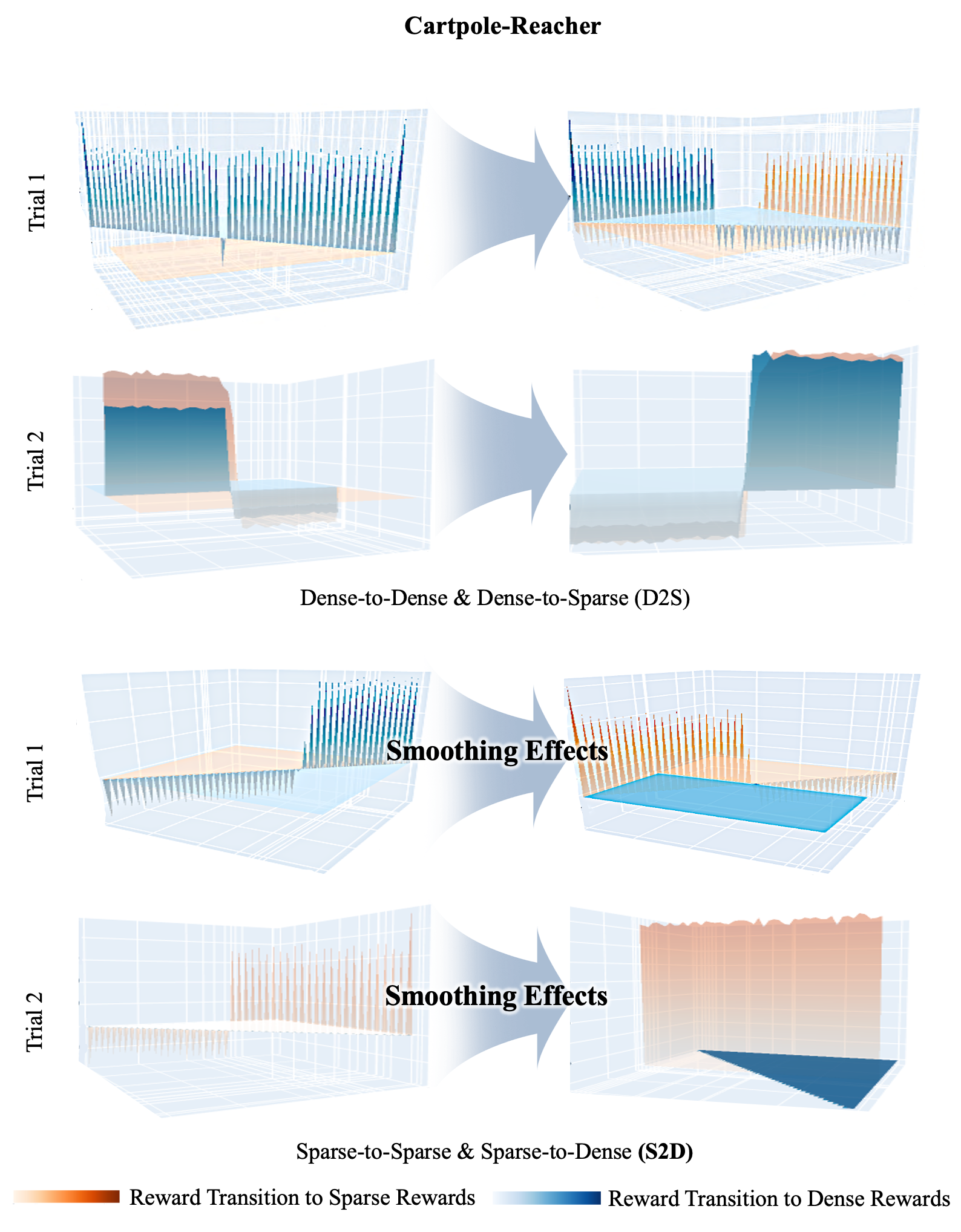}
\caption{This 3D visualization of the policy loss landscape for CartPole-Reacher illustrates that notable smoothing effects were predominantly seen with the Toddler-inspired S2D (\textcolor{blue}{sparse-to-dense}, $\mathscr{C}_2$, blue) transformation.}
\label{fig:cart_S2D}
\end{center}
\end{figure}

\begin{figure}[H]
\begin{center}
\includegraphics[width=\textwidth, height=0.8\textheight, keepaspectratio]{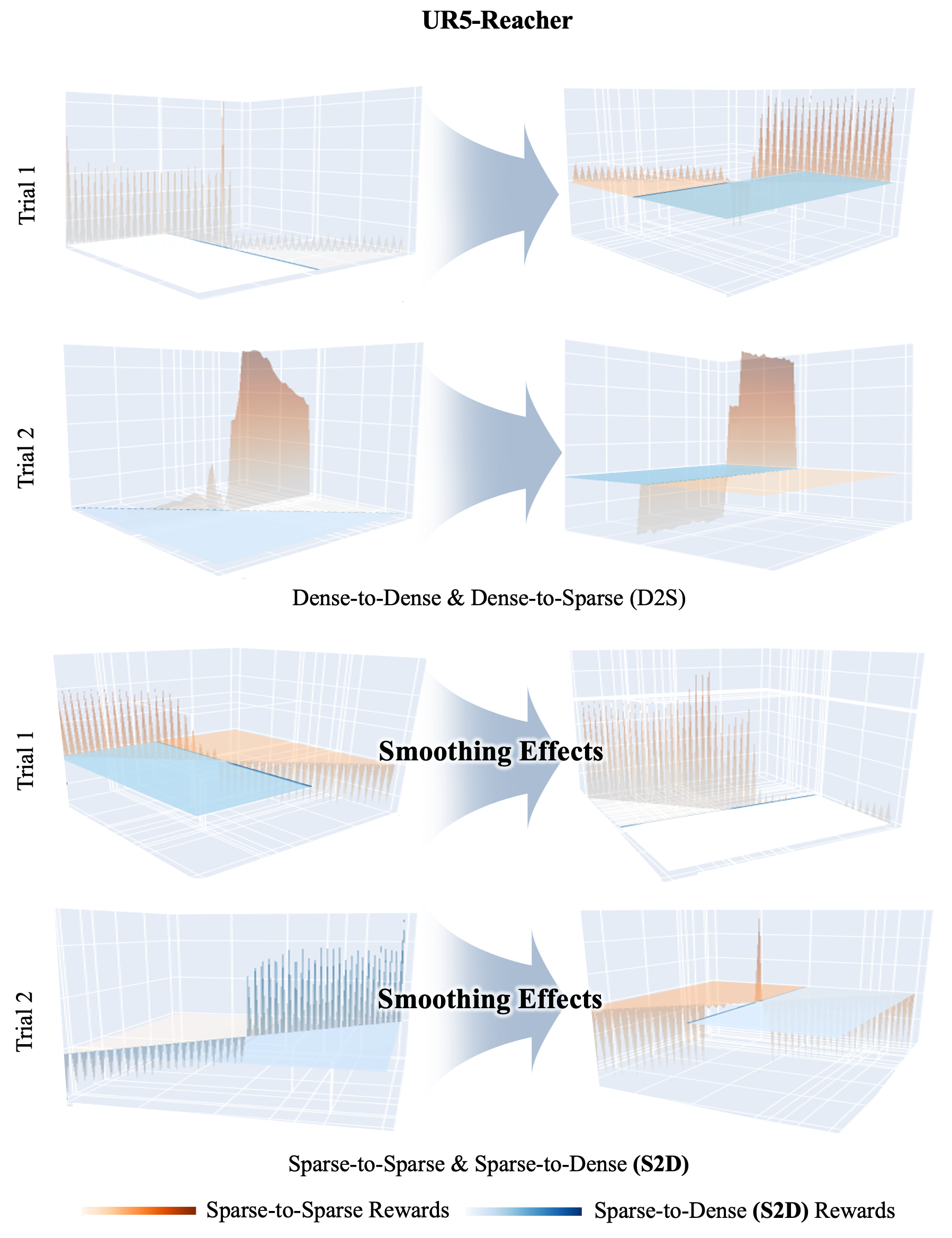}
\caption{This figure presents the 3D policy loss landscape for the UR5-Reacher task after various reward adjustments, such as sparse-to-dense (\textcolor{blue}{Toddler-inspired S2D}, $\mathscr{C}_3$, blue) and sparse-to-sparse (Only Sparse, red). Initially, the advantage of S2D over the Only Dense method becomes more pronounced over time, although both methods initially perform similarly. The Only Dense model displayed fewer lower depths of local minima after the reward transition, depending on seeds. However, the Toddler-inspired S2D method consistently exhibited significant smoothing effects more frequently, effectively reducing the depth of local minima and outperforming alternative baseline strategies.}
\label{fig:UR5_S2D}
\end{center}
\end{figure}

In contrast, the CartPole-Reacher and UR5-Reacher environments exhibit a more concentrated reward structure. The rewards are highly focused and localized, resulting in policy loss landscapes characterized by pronounced peaks, as illustrated in Figure~\ref{fig:cart_S2D} and \ref{fig:UR5_S2D} for CartPole-Reacher and UR5-Reacher, respectively. 

Through these visualizations, we gain a deeper understanding of the unique reward structures of various environments and how they shape policy loss landscapes, ultimately influencing the learning paths of agents after transitions.

\newpage
\section{Extensive 3D Visualizations for All Baseline Strategies}

\subsection{LunarLander-V2: Examining Dense-to-Sparse (D2S) \& Dense-to-Dense / Sparse-to-Dense (\textcolor{blue}{S2D}) \& Sparse-to-Sparse Transformations}

\newpage
\subsection{CartPole-Reacher: Assessing Dense-to-Sparse (D2S) \& Dense-to-Dense / Sparse-to-Dense (\textcolor{blue}{S2D}) \& Sparse-to-Sparse Shifts}

\newpage
\subsection{UR5-Reacher: Evaluating Dense-to-Sparse (D2S) \& Dense-to-Dense / Sparse-to-Dense (\textcolor{blue}{S2D}) \& Sparse-to-Sparse Dynamics}

\newpage

\newpage
\section{Section C: Additional Experiments of Toddler-Inspired S2D Reward Transition in Various Environments}
\label{app:exp}

\subsection{ViZDoom-FourObjects Navigation: Experiments on Reward Transition Timings} We developed additional environments using ViZDoom~\cite{kempka2016vizdoom}. These environments are slightly modified from those in \cite{kim2021goal}. In these environments, which we call \textit{ViZDoom-FourObjects} tasks, we focused on investigating the relationship between reward transition timings and learning.

\subsubsection{Task settings.}
\begin{figure}[!t]
    \centering
    \includegraphics[width=0.6\textwidth]{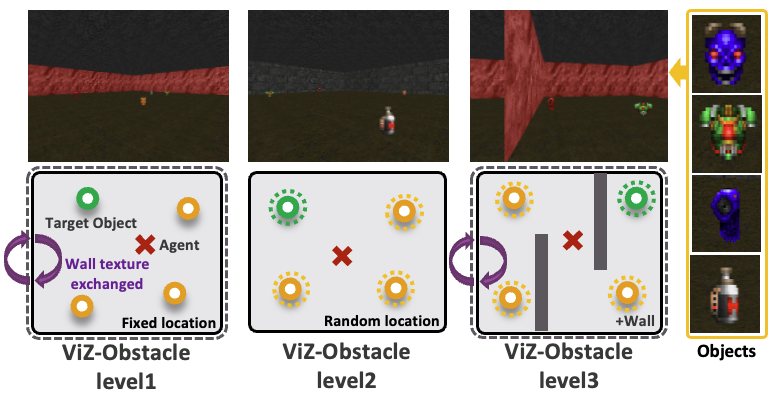}
    \caption{Overview of three ViZDoom-FourObjects environments: \textbf{Level1}, fixed object locations \& changing wall texture; \textbf{Level2}, random object locations; \textbf{Level3}, random object locations with changing wall textures and extra walls added.}
    \label{figure:vizEnvironment2}
\end{figure}

Four objects appear on the map, one of which is randomly selected as the target for the agent to locate. Each object can appear in two different colors or styles. The map is 700 by 700 units, and the agent starts in the middle for every episode. Figure~\ref{figure:vizEnvironment2} shows three ViZ-Level environments with increasing levels of complexity. Objects are positioned at a distance from the agent to keep the task challenging. Successfully reaching the goal object gives the agent a reward of 10.0, while touching a non-goal object results in a penalty of -1.0, ending the episode in either case. If the agent doesn’t reach any object within the time limit (25 steps for levels 1 and 2, or 37 steps for level 3), it receives a penalty of -0.1. To promote exploration, a reward of -0.01 is applied at each time step. We used the A3C algorithm~\cite{mnih2016asynchronous} for reinforcement learning, with averages and standard deviations calculated over three trials.

The main difference of these environments from ViZDoom-Seen and ViZDoom-Unseen is in the number of objects that are in the map, as well as the initial placement of objects and agent.
In the ViZDoom-Seen and Unseen, the agent is spawned at a corner of the map, while in ViZDoom-FourObjects environments, the agent is spawned near the center of the map.
The former environments have larger distance from the agent's initial position and the goal object's position in average, while the latter environments emphasize the need to distinguish between a wider diversity of objects.
In addition. the ViZ-Level3 environment adds extra walls within the map, which are not used in ViZDoom-Seen and Unseen.

\subsubsection{Settings on dense reward and curricula.}
The experiments on ViZDoom-FourObjects are unique in that three reward settings are covered, rather than two.
The sparse reward setting, referred to as \textit{Stage-1}, follows the reward scheme described above.
On top of this, \textit{Stage-2} provides an additional reward of 5.0 once the agent arrives within 200 unit distance from the goal object.
Lastly, \textit{Stage-3} provides an additional negative reward of -5.0 once the agent arrives within 200 unit distance of an object that is not the goal.
The tested three curricula settings are described in Figure~\ref{figure:Toddler-inspired S2D}, where $N$ is one million parameter updates.
Additionally, we tested the Only Dense setting, where only Stage-3 guidance is provided throughout the entire training.
We denote this setting as $\mathscr{C}_5$.

\subsubsection{Results on ViZDoom environments.}
\begin{figure}[t!]
    \centering
    \includegraphics[width= 0.95 \textwidth]{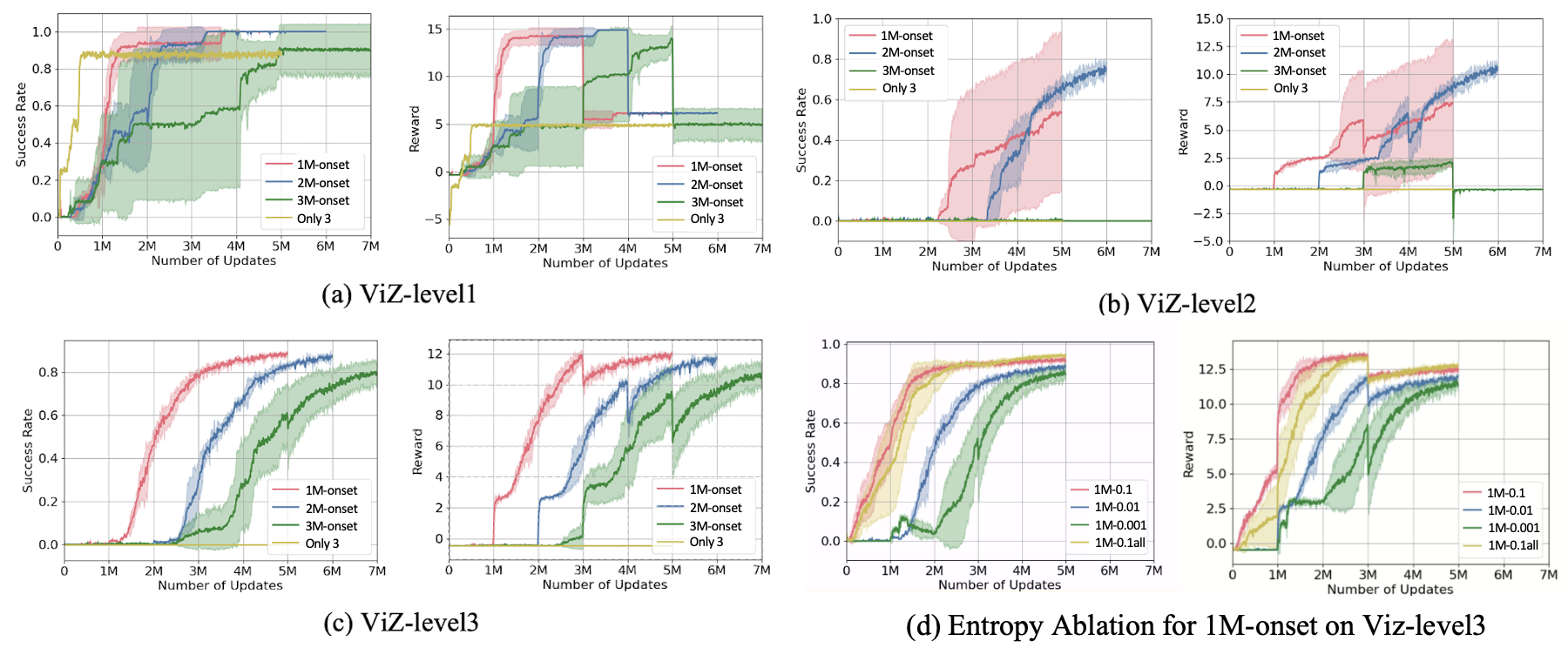}
    \caption{Comparison of different transition timings $(\mathscr{C}_1, \mathscr{C}_2, \mathscr{C}_3, \mathscr{C}_5$) according to three levels of ViZDoom-FourObjects. We use results from five trials.}
   \label{figure:vizResult}
\end{figure}

We demonstrate the impact of the three-stage transitions of Toddler-inspired S2D reward transition and examine the critical periods.
For each level of ViZDoom-FourObjects, we measure the agent's performance across three different stage transitions, with results displayed on Figure~\ref{figure:vizResult}.

Figure~\ref{figure:vizResult}-(a) displays the learning curves on ViZ-Level1. The agent reaches a perfect success rate (100\%) in the order of ($\mathscr{C}_1$) and ($\mathscr{C}_2$). ($\mathscr{C}_3$) shows the lowest success rate (92\%). The Only Dense model ($\mathscr{C}_5$) cannot even reach the lowest success rate of the Toddler-inspired S2D models (90.7\%).

Figure~\ref{figure:vizResult}-(b) displays the learning curves on ViZ-Level2.  $\mathscr{C}_2$ shows a superior success rate (78\%). The $\mathscr{C}_1$ also shows a moderate performance (57\%). In contrast, in $\mathscr{C}_3$ and $\mathscr{C}_5$, the agent cannot solve the task properly at all (0\%).

Lastly, Figure~\ref{figure:vizResult}-(c) displays the learning curves on ViZ-Level3, the most complex environment. Here, all models show a larger improvement according to stage-2,3 rewards. As shown in Figure~\ref{figure:vizResult}-(c), $\mathscr{C}_1$ (90\%), $\mathscr{C}_2$ (83\%) and $\mathscr{C}_3$ (78\%) exhibit best to worst performances in order.

\subsubsection{Overall analysis on ViZDoom-FourObjects environments.}
We observe vast performance improvements after first transition from Stage-1 to Stage-2 rewards at the \textit{1M point} in ViZ-Level1 and Level3, particularly with the best performing models of ViZ-Level1 $\mathscr{C}_1$ and Level3 ($\mathscr{C}_1$) respectively. In the case of Viz-Level2, $\mathscr{C}_2$, where stage transition occurs at 2M, has shown the most outstanding performance, while $\mathscr{C}_3$ cannot learn the task at all.
Therefore, we observe that there is the \textit{appropriate} timing of stage transition within Toddler-inspired S2D, which leads to the steepest performance improvement in these visual navigation tasks. 
Especially, the initial stage transition, whether at 1M updates for ViZ-Level1 and Level3 or at 2M updates for ViZ-Level2, highlights a crucial timing for reward transition strategies. We examined the importance of pinpointing this optimal phase for moving from sparse to dense rewards, taking cues from toddler developmental stages.

\subsubsection{Importance of early free exploration.}
As additional ablative experiments, we varied the entropy prior to the transition from Stage-1 to Stage-2 reward.
The prior entropy was set to \{0.1, 0.01, 0.001\}, while the entropy afterwards was fixed to 0.01.
As shown in Figure~\ref{figure:vizResult}-(d), we found that an appropriate entropy term of 0.1 is more crucial for the Toddler-inspired S2D reward transition than 0.01 (blue) and 0.001 (green), indicating the importance of free exploration at the early stage of learning.

\newpage
\newpage
\subsection{Shelf Delivery Tasks in RWARE: Experiments Using a Three-Stage Guidance Approach with Suboptimal Rewards}

\subsubsection{Task Configuration.}
\begin{figure*}[t!]
    \centering
    \includegraphics[width=0.9\textwidth]{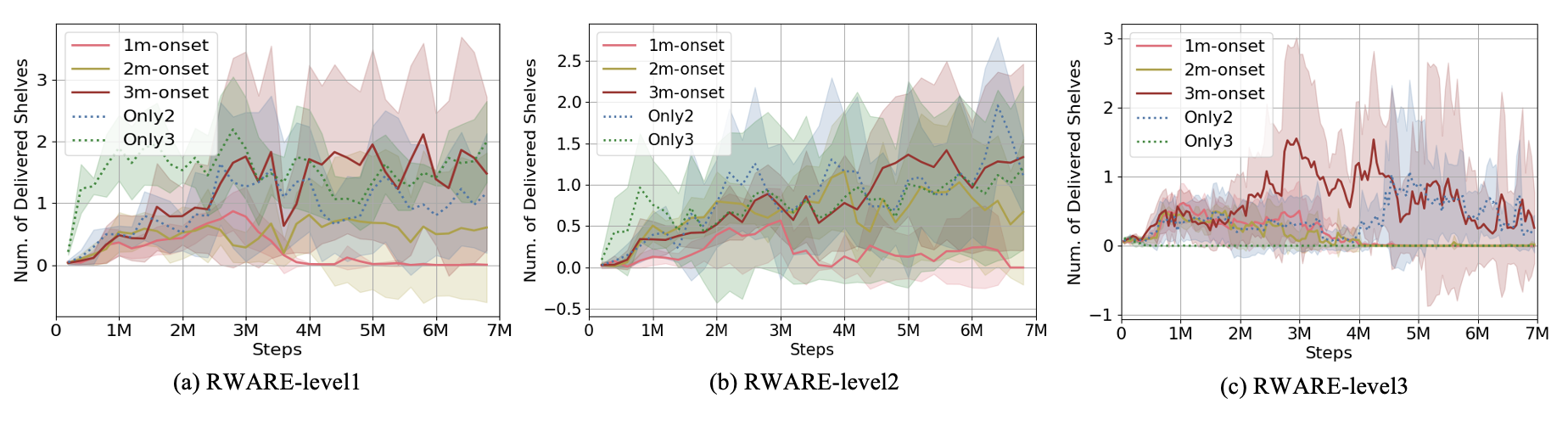}
    \caption{Performance metrics in RWARE tasks. The vertical axis measures the average number of shelves successfully delivered per episode, while the horizontal axis records the time steps. Solid lines indicate the performance of agents utilizing Toddler-inspired S2D transitions $(\mathscr{C}_1)$, $(\mathscr{C}_2)$, $(\mathscr{C}_3)$, whereas dotted lines depict the outcomes of agents using only dense rewards $(\mathscr{C}_4)$. The mean and standard deviation for each trial, conducted over five attempts, are represented by lines and shaded areas. Notably, the $(\mathscr{C}_3)$ agent demonstrates superior learning curves across all scenarios, even in situations where Stage-3 guidance proves less useful, as seen in Level 3. The large standard deviations underscore the inherent challenges in RWARE tasks with multiple subgoals.}
   \label{fig:resultG2}
\end{figure*}

Figures~\ref{fig:resultG2} and \ref{Rwaretrajectory} showcase our expanded tests within the RWARE grid-world environments, which utilize a discrete state-action space. We have tailored RWARE \cite{papoudakis2021benchmarking} for single-agent operations, involving a mobile agent navigating through rows of shelves (\textcolor{blue}{blue}), some randomly tagged as "requested" (\textcolor{red}{red}) for delivery. The agent's actions include \{\textit{MoveForward, TurnLeft, TurnRight, Load/Unload, Noop}\}. It can only sense the tiles in a 3x3 grid around its position. The agent's objective is to move requested shelves to a goal location (\textcolor[HTML]{3d8c40}{green}) and then return them, completing a series of subgoals, such as reaching, transporting, and restoring the shelves.

Three levels of difficulty were set: Level 1 includes 8 shelves, 3 requested; Level 2 has 10 shelves, 3 requested, leading to a sparser reward structure; Level 3 features 32 shelves with 16 requested, greatly increasing the complexity of exploration. Figure~\ref{Rwaretrajectory} displays these setups.

\subsubsection{Guidance Strategy.}
Rewards were structured across three stages reflecting the subtasks: Stage-1 gives a +1.0 reward for delivering a requested shelf to the destination; Stage-2 rewards both delivering and returning the shelf; Stage-3 adds a bonus for picking up a requested shelf.

\subsubsection{Results in RWARE Experiments.}
This investigation examines how different Toddler-inspired S2D reward schemes affect learning tasks in RWARE, focusing on \textit{subgoal} sequences to observe suboptimal outcomes near transition phases. We assessed Toddler-inspired S2D transitions at various stages ($\mathscr{C}_1$, $\mathscr{C}_2$, $\mathscr{C}_3$) and only dense rewards at Stage-2 ($\mathscr{C}_4$) or Stage-3 ($\mathscr{C}_5$). Transition timings are as described in Figure~\ref{figure:Toddler-inspired S2D}, with $N$ set to 1 million (1M) time steps. PPO \cite{Schulman2017ProximalPO} was employed as the learning algorithm, and outcomes were averaged over five trials. Here are the main results:

\begin{itemize}
\item \textbf{Level 1.} (Figure \ref{fig:resultG2}-(a)).
Among Toddler-inspired S2D agents, $\mathscr{C}_3$ showed superior performance, while $\mathscr{C}_1$ did not manage to deliver shelves beyond the 4M time step. Although Only 2 ($\mathscr{C}_4$) and Only 3 ($\mathscr{C}_5$) achieved impressive results, the $\mathscr{C}_3$ agent was on par or better.

\item \textbf{Level 2.} (Figure \ref{fig:resultG2}-(b)).
In this scenario, $\mathscr{C}_3$ again led the performance, with $\mathscr{C}_2$ following closely. The performances of Only 2 ($\mathscr{C}_4$) and Only 3 ($\mathscr{C}_5$) were comparable to $\mathscr{C}_3$.

\item \textbf{Level 3.} (Figure \ref{fig:resultG2}-(c)).
In this challenging environment, $\mathscr{C}_3$ excelled. Stage-3 guidance was not particularly beneficial, as seen with the Only 3 agent ($\mathscr{C}_4$), but $\mathscr{C}_3$ maintained robustness against less advantageous reward settings.
\end{itemize}

\subsubsection{Comprehensive Analysis of RWARE Tasks.}
\begin{figure}[t!]
\centering
    \includegraphics[width=0.9\columnwidth]{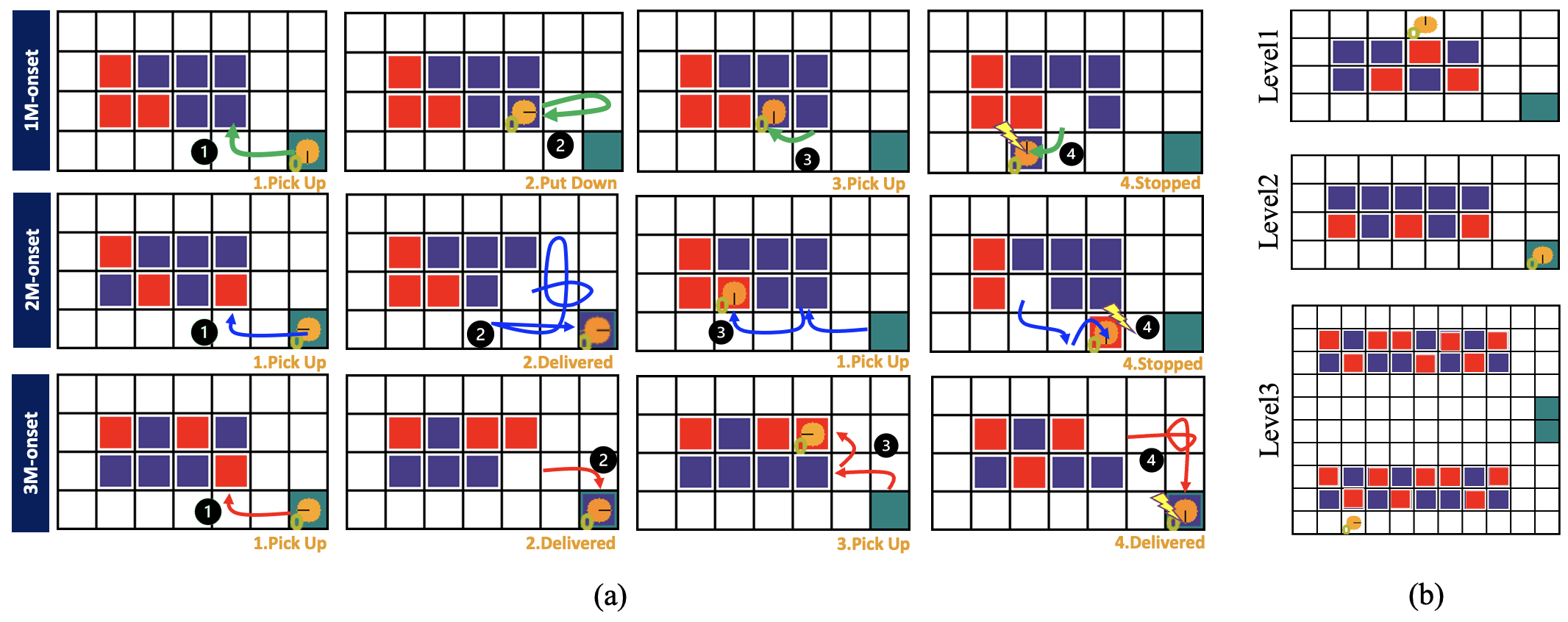}
    \caption{(a) Visual representation of agent trajectories in RWARE-Level1. (b) The three different levels of RWARE environments.}
   \label{Rwaretrajectory}
\end{figure}

In Figure~\ref{Rwaretrajectory}, we illustrate the movement paths of various agents in RWARE-Level1. Goals, requested shelves, and unrequested shelves are represented by green, red, and blue squares. The agent starts where the goal is located (green). The $\mathscr{C}_1$ agent mistakenly selected incorrect shelves twice before halting. The $\mathscr{C}_2$ agent managed to deliver correctly initially but struggled with efficient navigation afterward. The $\mathscr{C}_3$ agent, however, delivered shelves efficiently and minimized subgoal failures, emphasizing the importance of well-timed reward transitions.

\subsubsection{Significance of Early Exploration.}
$\mathscr{C}_1$ and $\mathscr{C}_2$ underperformed when initial stage transitions occurred prematurely, before 3M training steps, as seen in Figure~\ref{fig:resultG2}. This highlights the necessity for an adequate exploration period (Stage-1) with sparse rewards for effective learning. Despite having rich rewards, the Only 3 agent ($\mathscr{C}_5$) failed completely in Level 3. Conversely, the $\mathscr{C}_3$ agent showed resilience to less beneficial rewards during initial stage transitions in the Toddler-inspired S2D framework.

\subsubsection{Study Limitations.}
This study's main goal was to assess how the timing of reward transitions impacts performance metrics in tasks featuring subgoals. The findings detail how these transitions influence overall task outcomes. However, determining an optimal reward shaping strategy for tasks with subgoals remains a challenge. This presents a key area for further investigation, focusing on developing more refined reward shaping techniques, including potential-based methods within this specific context, to enhance the effectiveness of toddler-inspired reward transitions and improve performance outcomes.


\subsection{Gridworld: Add-on Algorithms Experiments on 3D Loss Landscape}
\label{sec:gridworld}
\begin{figure}[!t]
\centering
    \includegraphics[width=0.9\textwidth]{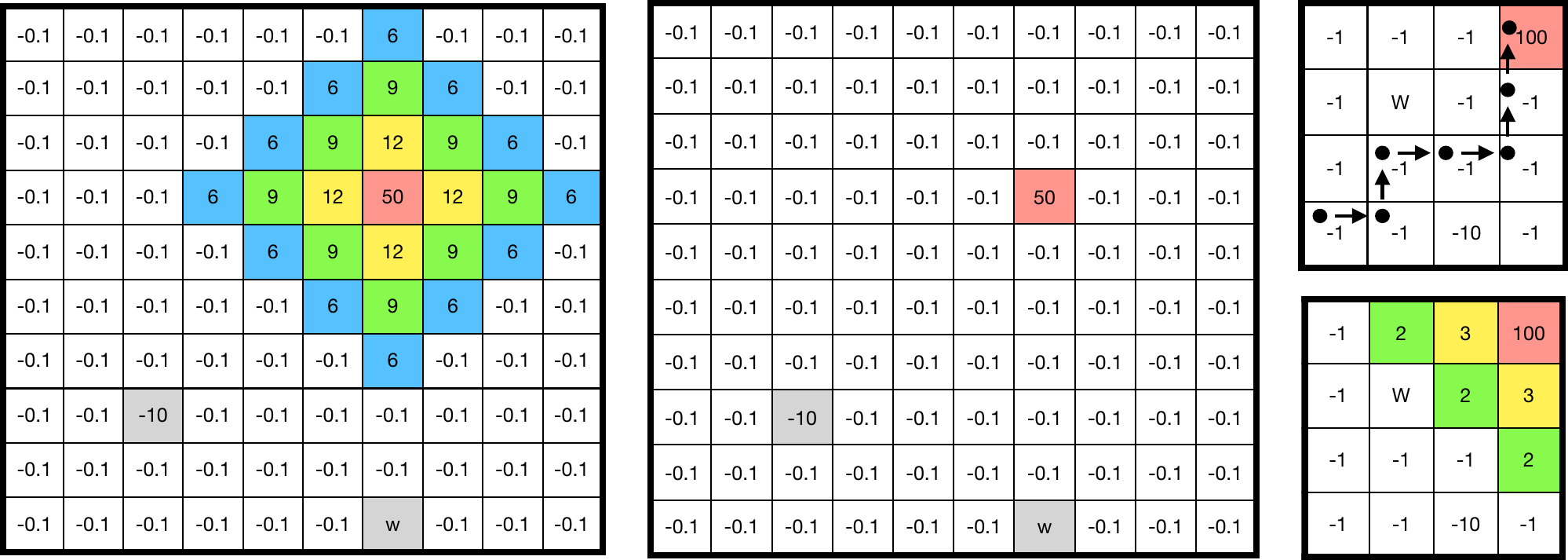}
    \caption{Gridworld-navigation task. \textbf{Left}: $10\times 10$ environment with potential-based dense rewards using \textit{PPO}. \textbf{Center}: $10\times 10$ environment with sparse rewards using \textit{PPO}. \textbf{Right-Top}: $4\times 4$ environment with sparse rewards using \textit{DQN}. \textbf{Right-Bottom}: $4\times 4$ environment with potential-based dense rewards using \textit{DQN}.
}
   \label{fig:dense_sparse}
\end{figure}

To explore the impact of reward transitions on policy learning, we conducted 3D visualizations of the policy loss landscape using the Gridworld environment. Gridworld’s simplicity provides a controlled setting with fewer variables, allowing us to clearly observe the effects of the S2D reward transition compared to baseline methods. In our main analysis, we employed the SAC algorithm \cite{haarnoja2018soft}, but to ensure our observations were not limited to a single algorithm, we extended our experiments to include DQN \cite{dqn2} and PPO \cite{Schulman2017ProximalPO}.

\subsubsection{Environment Setup}
The experimental setup features a gridworld environment, as shown in Figure~\ref{fig:dense_sparse}. The agent can move in four directions: up, down, left, and right, and receives a living penalty of $-0.1$ to encourage exploration. We performed experiments in two scenarios: (1) a fixed-goal $4\times 4$ environment using DQN \cite{dqn2}, and (2) a random-goal $10\times 10$ environment using PPO \cite{Schulman2017ProximalPO}. To achieve optimal stage transition, we set $T=200$ for the fixed goal over 1000 steps and $T=5000$ for the random goal over 100,000 steps. The neural network architecture comprises three fully connected layers with ReLU activation, and the batch size is set to 128. For PPO, updates were made every 2 episodes. The Cross-Density Visualizer strategy was employed for visualization.

\subsubsection{Results of Loss Landscape in DQN and PPO Algorithms}
\label{sec:strategy}

\begin{figure}[!hb]
\centering
    \includegraphics[width=\textwidth, height=0.75\textheight, keepaspectratio]{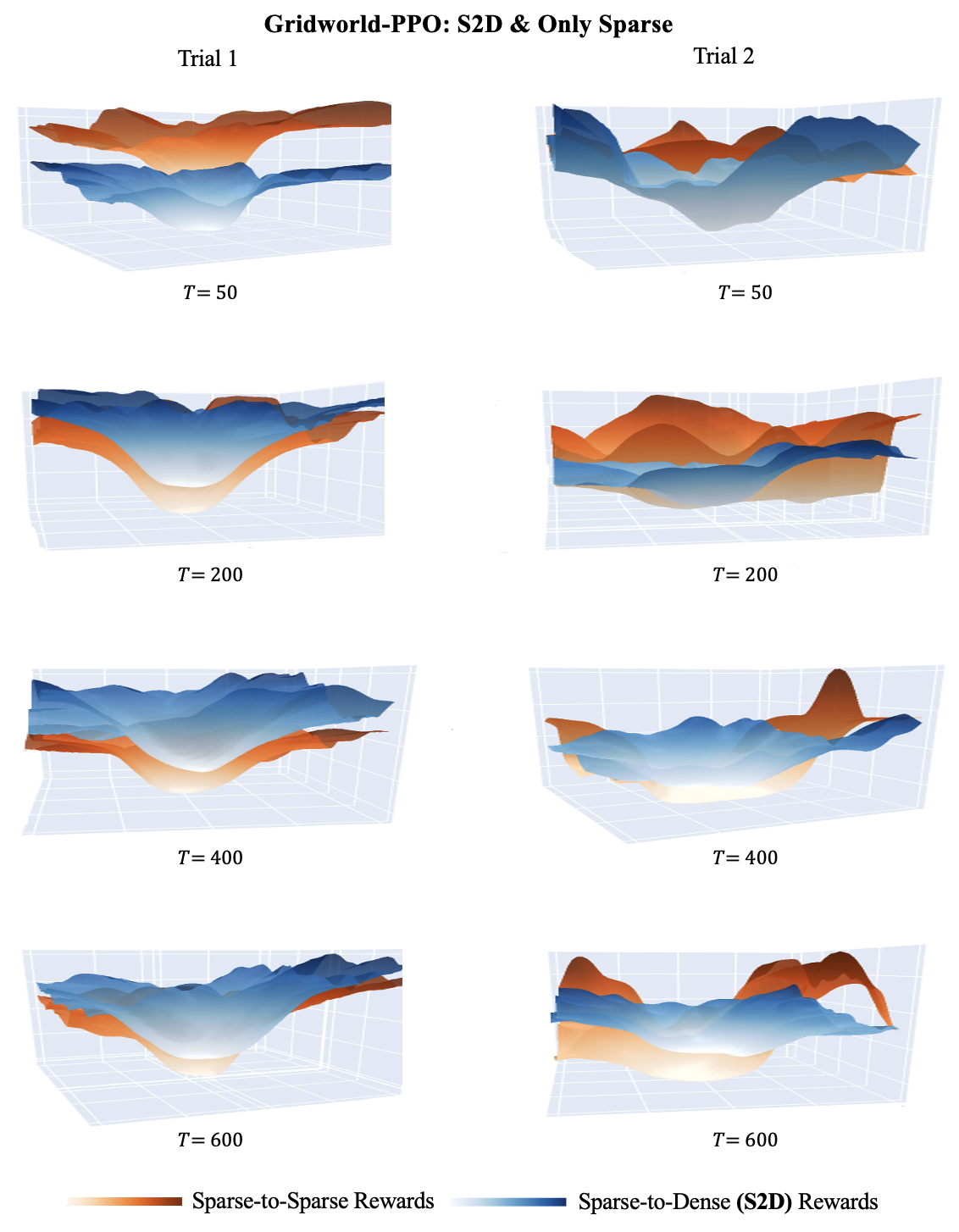}
    \caption{This visualization showcases the 3D policy loss landscape post-transition from sparse-to-dense (\textcolor{blue}{Toddler-inspired S2D}, $\mathscr{C}_3$, blue) and sparse-to-sparse (Only Sparse, red) reward schemes. The Toddler-inspired S2D transition in the PPO algorithm exhibits a pronounced smoothing effect, notably reducing the depth of local minima.}
   \label{fig:girdppos2d}
\end{figure}

\begin{figure}[!hb]
\centering
    \includegraphics[width=\textwidth, height=0.8\textheight, keepaspectratio]{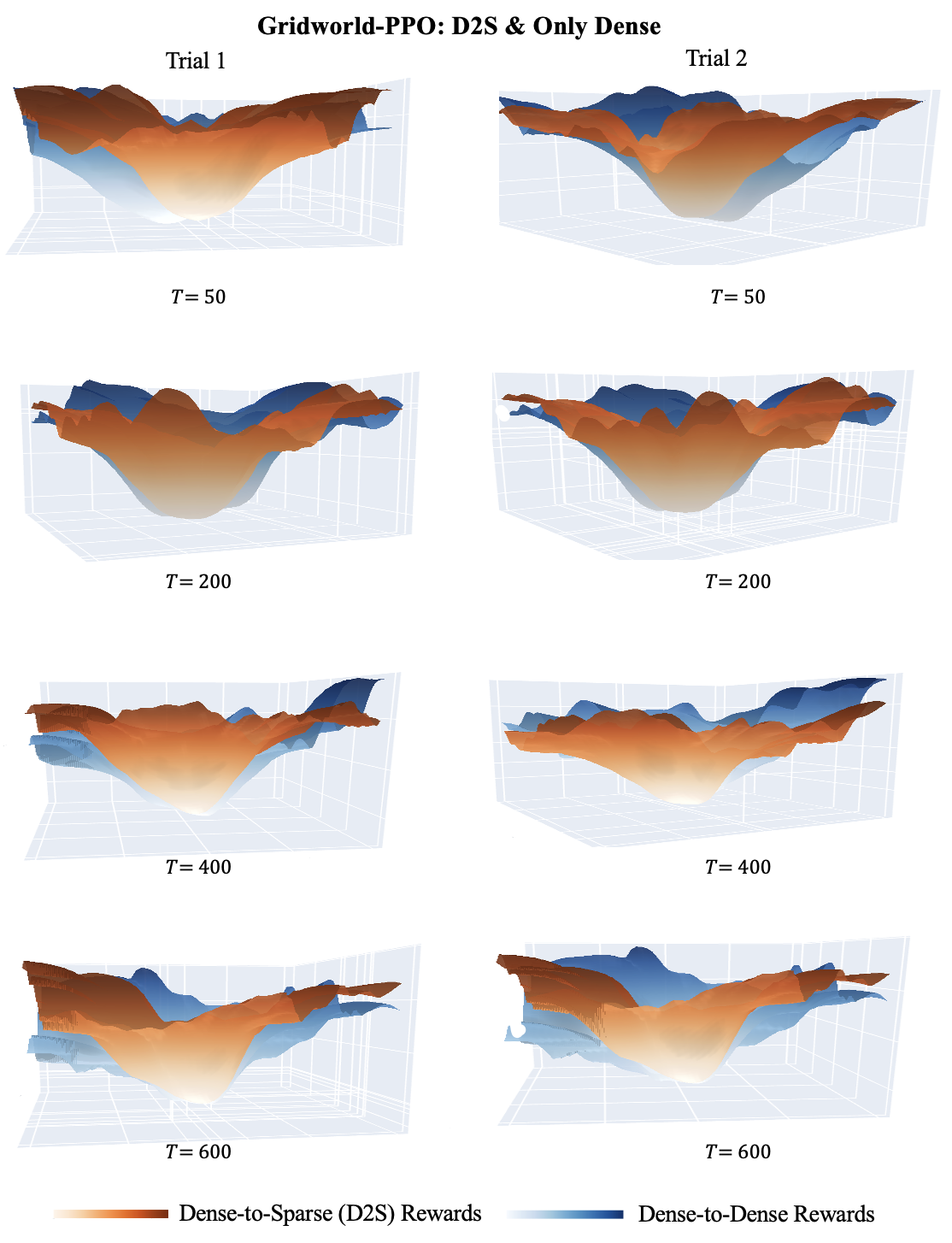}
    \caption{This figure presents the 3D policy loss landscape for dense-to-sparse (D2S, $\mathscr{C}_3$, red) and dense-to-dense (Only Dense, blue) transitions. As observed in Appendix Section B, these configurations show minimal smoothing effects, even with increased updates, indicating less adaptability in the Gridworld-PPO setup.}
   \label{fig:girdppod2s}
\end{figure}

\begin{figure}[!hb]
\centering
    \includegraphics[width=1\textwidth]{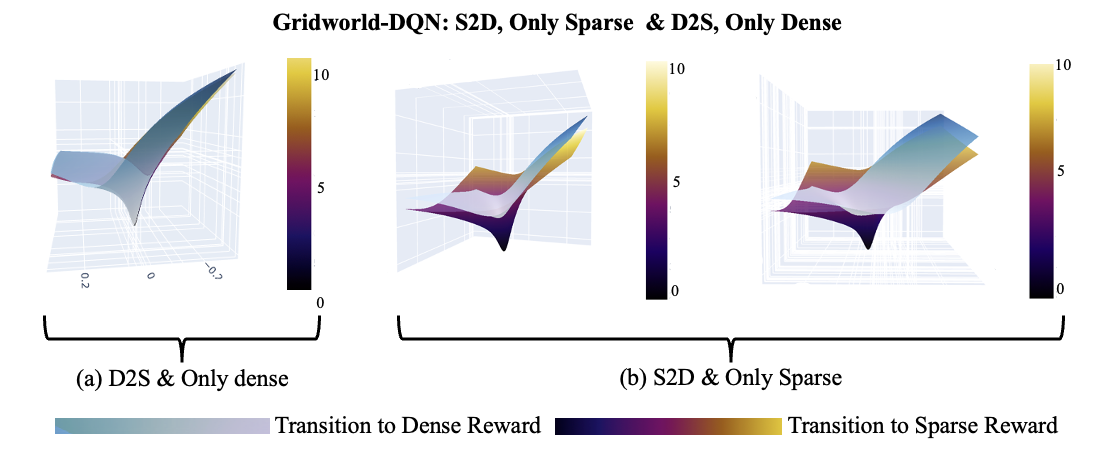}
    \caption{This illustration captures the 3D Q-value loss landscape across different reward transitions, including Toddler-inspired S2D and baseline approaches. Unlike the baseline methods (D2S, Only Dense, Only Sparse), which lack significant smoothing, the \textcolor{blue}{Toddler-inspired S2D} transition in panel (b) demonstrates a clear smoothing effect by effectively reducing the depth of local minima through the adoption of potential-based dense rewards.}
   \label{fig:dense_sparse2}
\end{figure}

We analyzed the policy loss landscape using the PPO algorithm and the Q-function loss landscape using the DQN algorithm. While DQN focuses on learning Q-values, examining its loss landscape offers insights into how different reward schemes affect learning dynamics. In both Gridworld-DQN and PPO scenarios, the Toddler-inspired S2D reward transition demonstrated a noticeable smoothing effect on the loss landscape, outperforming other baseline methods. This effect is highlighted in Figures \ref{fig:girdppos2d}, \ref{fig:girdppod2s}, and \ref{fig:dense_sparse2}(b), where it achieved superior performance metrics. However, as shown in Figures \ref{fig:girdppod2s} and \ref{fig:dense_sparse2}-(a), there is little distinction between dense and sparse landscapes as updates increase. Our findings suggest that transitioning from sparse to dense rewards using the Toddler-inspired S2D method effectively reduces local minima depth, surpassing both dense-to-sparse (D2S) and exclusively dense approaches.

\clearpage
\subsection{In-Depth 3D Loss Landscape Analysis for Gridworld-PPO and DQN Algorithms}

\subsubsection{Gridworld-PPO: Exploring Sparse-to-Dense (\textcolor{blue}{S2D}) \& Sparse-to-Sparse Transitions}
\clearpage

\subsubsection{Gridworld-PPO: Analyzing Dense-to-Sparse (D2S) \& Dense-to-Dense Shifts}
\clearpage

\subsubsection{Gridworld-DQN: Comparing \textcolor{blue}{S2D}, Only Sparse, D2S, and Only Dense Approaches}



\end{document}